\documentclass[10pt,twocolumn,letterpaper]{article}

\usepackage[pagenumbers]{cvpr} % To force page numbers, e.g. for an arXiv version

\usepackage[accsupp]{axessibility}
\usepackage{graphicx}
\usepackage{amsmath}
\usepackage{amssymb}
\usepackage{booktabs}
\usepackage{multirow}
\usepackage{makecell}
\usepackage{float}

\newcommand{\highlight}[1]{{\color{black}#1}}

\usepackage[pagebackref,breaklinks,colorlinks]{hyperref}

\usepackage[capitalize]{cleveref}
\crefname{section}{Section}{Sections}
\Crefname{section}{Section}{Sections}
\Crefname{table}{Table}{Tables}
\crefname{table}{Table}{Tables}

\usepackage{enumitem}

\def\cvprPaperID{5946} % *** Enter the CVPR Paper ID here
\def\confName{CVPR}
\def\confYear{2023}

\begin{document}

\title{Toward Verifiable and Reproducible Human Evaluation \\for Text-to-Image Generation}

\author{Mayu Otani$^{1}$ \and
Riku Togashi$^{1}$ \and
Yu Sawai$^{1}$ \and
Ryosuke Ishigami$^{1}$ \and
Yuta Nakashima$^{2}$ \and
Esa Rahtu$^{3}$ \and
Janne Heikkil\"{a}$^{4}$ \and
Shin'ichi Satoh$^{1}$ \and \\
$^{1}$CyberAgent, Inc. \qquad
$^{2}$Osaka University \qquad
$^{3}$Tampere University \qquad
$^{4}$University of Oulu
}

\maketitle

%%%%%%%%% ABSTRACT
\begin{abstract}

Human evaluation is critical for validating the performance of text-to-image generative models, as this highly cognitive process requires deep comprehension of text and images. 
However, our survey of 37 recent papers reveals that many works rely solely on automatic measures (\eg, FID) or perform poorly described human evaluations that are not reliable or repeatable.
This paper proposes a standardized and well-defined human evaluation protocol to facilitate verifiable and reproducible human evaluation in future works.  
In our pilot data collection, we experimentally show that the current automatic measures are incompatible with human perception in evaluating the performance of the text-to-image generation results. Furthermore, we provide insights for designing human evaluation experiments reliably and conclusively. 
Finally, we make several resources publicly available to the community to facilitate easy and fast implementations. 

\end{abstract}

%%%%%%%%% BODY TEXT
\section{Introduction}
\label{sec:intro}
%With the great growth of interest in text-to-image synthesis, many papers have been published with new models. 
Text-to-image synthesis has seen substantial development in recent years. Several new models have been introduced with remarkable results. The majority of the works validate their models using automatic measures, such as FID \cite{fid_neurips_2017} and recently proposed CLIPScore \cite{hessel-etal-2021-clipscore}, even though many papers point out problems with these measures. The most popular measure, FID, is criticized for misalignment with human perception \cite{ding_cogview2_2022}. For example, image resizing and compression hardly degrade the perceptual quality but induce high variations in the FID score \cite{aliased_resizing_2022}, while CLIPScore can inflate for a model trained to optimize text-to-image alignment in the CLIP space \cite{glide_icml_2022}.

This empirical evidence of the misalignment of the automatic measures motivates human evaluation of perceived quality. However, according to our study of 37 recent papers, the current practices in human evaluation face significant challenges in reliability and reproducibility. 
%However, current practices of human evaluation are experiencing fundamental challenges in reproducibility according to our survey of 37 recent publications. 
%There are mainly two problematic practices risking the reliability of human evaluation.
We mainly identified the following two problematic practices.
Firstly, evaluation protocols vary significantly from one paper to another. For example, some employ relative evaluation by simultaneously showing annotators two or more samples for comparison, and others collect scores of individual samples based on certain criteria. 
Secondly, important details of experimental configurations and collected data are often omitted. For example, the number of annotators who rate each sample is mostly missing, and sometimes even the number of assessed samples is not reported. These inconsistencies in the evaluation protocol make future analysis almost impossible. We also find that recent papers do not analyze the quality of the collected human annotations. Therefore, we cannot assess how reliable the evaluation result is. It is also difficult to know which is a good way to evaluate text-to-image synthesis among various evaluation protocols.

\begin{figure}[t!]
    \centering
    \includegraphics[width=0.9\linewidth,clip]{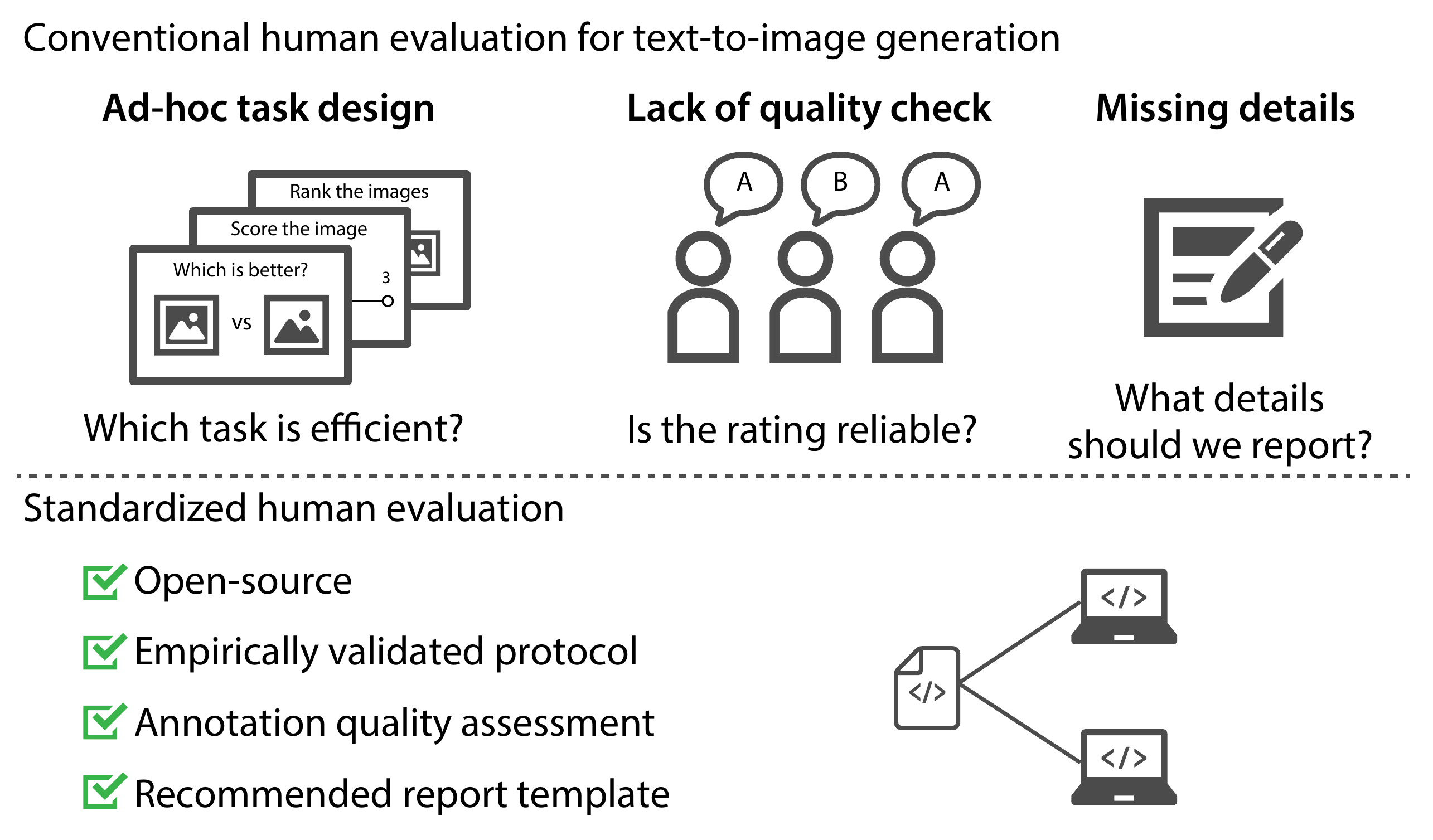}
    \caption{\highlight{Conventionally, researchers have used different protocols for human evaluation, and setup details are often unclear. We aim to build a standardized human evaluation.}}
    \label{fig:teaser}
\end{figure}
% 画像ももうちょい変わったことを強調

The natural language generation (NLG) community has extensively explored human evaluation.
Human evaluation is typically done in a crowdsourcing platform, and there are many well-known practices. Yet, quality control is an open challenge \cite{perils-2021}.
Recently, a platform for benchmarking multiple NLG tasks was launched \cite{genie-2022}.
The platform offers a web form where researchers can submit their model predictions. The platform automatically enqueues a human evaluation task on AMT, allowing a fair comparison for its users.

% Our goal is to uncover these problems of human evaluation in text-to-image synthesis and build a standard protocol for better reproducibility. 
We address the lack of a standardized evaluation protocol in text-to-image generation.
To this end, we carefully design an evaluation protocol using crowdsourcing and empirically validate the protocol.
% We publish our implementation to facilitate reliable and transparent human evaluation. 
We also provide recommendations for reporting a configuration and evaluation result.

% The experiments validate our design choices by pilot data collection. 
We evaluate state-of-the-art generative models with our protocol and provide an in-depth analysis of collected human ratings. We also investigate automatic measures, \ie, FID and CLIPScore, by checking the agreement between the measures and human evaluation. The results reveal the critical limitations of these two automatic measures. 

\paragraph{Findings and resources}

Our main findings can be summarized as follows:

\begin{itemize}
\setlength{\parskip}{0cm}
\setlength{\itemsep}{1mm}
    \item \textbf{Reliability of prior human evaluation is questionable.} We provide insights about the required numbers of prompts and human ratings to be conclusive. 
    \item \textbf{FID is inconsistent with human perception.} This is already known at least empirically, and our experiment supports it.
    % \item \textbf{CLIPScore suffers from adversarial effects.} Models optimized in the CLIP space yield high scores that are incompatible with human perception.
    \item \textbf{CLIPScore is already saturated.} State-of-the-art generative models are already on par with authentic images in terms of CLIPScores.
\end{itemize}

These findings motivate us to build a standardized protocol for human evaluation for better verifiability and reproducibility, facilitating to draw reliable conclusions. For continuous development, we open-source the following resources for the community. 
\begin{itemize}
\setlength{\parskip}{0cm}
\setlength{\itemsep}{1mm}
    \item Implementation of human evaluation on a crowdsourcing platform, \ie, Amazon Mechanical Turk (AMT)\footnote{https://www.mturk.com/}, which allows researchers to evaluate their generative models with a standardized protocol.
    \item Template for reporting human evaluation results. This template will enhance their transparency.
    \item Human ratings by our protocol on multiple datasets: MS-COCO \cite{ms-coco}, DrawBench \cite{imagen_drawbench}, and PartiPrompts \cite{parti_2022_arxiv}. These per-image ratings, and not only their statistics, will facilitate designing automatic measures.
\end{itemize}

\section{Related work}
% \textbf{text-to-image synthesis}
% text-to-image synthesis
% Large-scale TTI published a lot

\textbf{Human evaluation}
Evaluation of generative models, \eg, for perceptual and linguistic data, inherently involves human perception and understanding, so human evaluation is inevitable. Nevertheless, there are no established evaluation practices, and researchers have been using different protocols \cite{ding_cogview2_2022,maharana_storydall-e_2022,stackgan_2017_iccv,parti_2022_arxiv,yan_trace_2022}. 
To address challenges in human evaluation due to ad-hoc practices, prior studies offer shared human evaluation protocols, such as the one for unconditional image synthesis \cite{hype-neurips2019}.
NLG community provides in-depth analysis of challenges in human evaluation on story generation \cite{perils-2021}. To facilitate reliable model comparison based on human evaluation, a platform hosting human evaluation of multiple language generation tasks is proposed \cite{genie-2022}.
Inspired by these works, we aim to develop a shared evaluation protocol for text-to-image generation.

Some literature reviews have summarized challenges in crowdsourcing \cite{shape-solution-2020,mturk_research}.
They conclude that guaranteeing annotators' reliability is the main challenge.
Annotators have a strong incentive to maximize their own monetary returns in the shortest possible period. As a result, they do not often pay enough attention to the tasks.
The reviews offer actionable techniques to alleviate such problems.

\textbf{Automatic evaluation}
The community has granted automatic measures, 
%Most prior studies on unconditional image generation evaluate a model's output quality using automatic measures, 
such as Inception Score (IS) \cite{is_neurips_2016}, Fre\'chet Inception Distance (FID) \cite{fid_neurips_2017}, and Precision-Recall \cite{prec_recall_neurips_2019}, as additional options for evaluation.
IS evaluates whether generated images have identifiable objects and diversity using the output of the Inception model \cite{inceptionnet}.
%IS relies on visual concepts that the Inception model learned and assumes that there is a single visual concept in each image. IS thus fails to handle unseen or multiple visual concepts in a single image.
FID and Precision-Recall evaluate the discrepancy between distributions of real and generated images.

Evaluation of text-to-image generative models often leverages R-precision \cite{rprec_attngan} and CLIPScore \cite{hessel-etal-2021-clipscore}.
CLIPScore has recently been proposed to evaluate image-text alignment in image captioning, which is diverted to text-to-image generation. 
Concept detectors can also be used to assess if a predefined set of concepts in the prompt are detected in the generated image \cite{liu2022compositional}.
Meanwhile, many prior works pointed out the limitations of these automatic measures \cite{doggan-kdd21,kid_iclr18,cafd2018,aliased_resizing_2022,ding_cogview2_2022,faithful_precision_recall_icml22}.
Image processing operations, such as resizing and compression, often exemplify the misalignment problem of FID with human perception \cite{aliased_resizing_2022,cafd2018}, which is empirically confirmed using human ratings \cite{ding_cogview2_2022}.

\section{Review: Human evaluation in text-to-image}
\label{sec:metaeval}

We surveyed 37 recent text-to-image generation papers\footnote{We collected papers by querying ``text to image'' in CVF conferences, including CVPR, ICCV, and ECCV from 2017 to 2022. We also add recent successors. The full list of papers is in the supplementary material.}, and reviewed how they use and report human evaluation. The details of the surveyed papers are in the supplementary material. The following summarizes our findings. 

\textbf{Human evaluation vs.~automatic measures}
Only 20 out of 37 papers provide human evaluation, which means that 17 works rely solely on automatic measures. Such measures are found to be inconsistent with human perception, therefore, should not be used as the only measure. 

\highlight{\textbf{Number of samples and ratings}}
Among 20 papers, 18 papers report the number of samples.
However, only four disclose the number of ratings per sample even though the evaluation of generated images is highly subjective, and large discrepancies are expected in ratings.
Moreover, crowdsourcing often suffers from noise and requires multiple annotations for each sample to be conclusive.

\textbf{Evaluation criteria}
The overall quality of generated images and relevance to text prompts are major concerns in human evaluation; 18 papers assess overall quality and 14 papers assess relevance to text. Others include correctness of object locations \cite{yan_trace_2022} and consistency of multiple image generation \cite{maharana_storydall-e_2022}. This implies that some papers only evaluate a single aspect of generated images. 

\textbf{Rating methods} We identify three different methods to collect ratings.
Ten papers adopt a comparative approach by choosing the best among two or more samples, while nine works use comparative judgment but require ranking multiple samples.
%Ten papers adopt a comparative approach by choosing the best among two or more samples. Nine papers still use comparative judgment but require ranking multiple samples. 
Three papers use a 5- or 3-point Likert scale for rating. There are wide discrepancies in the way of conducting human evaluation in different papers. However, their validity is rarely discussed.

\textbf{Annotation quality assessment} We find a problematic practice of not reporting the quality of annotations. A typical metric is an inter-annotator agreement (IAA), such as Cohen's $\kappa$ and Krippendorff's $\alpha$ \cite{klaus1980content}. No paper reports IAA, which raises a concern about the reliability of results.

\textbf{Sample size}
Many papers use less than 100 samples for each model for human evaluation \cite{ding_cogview_2021,ding_cogview2_2022,liang_cpgan_2020,li_stylet2i_2022,avidan_tise_2022}.
Such a small sample size occasionally leads to different conclusions.
Our experiment on COCO in \cref{sec:sample_size} demonstrates that more than 500 samples are necessary for a stable ranking of competing models; otherwise, the ranking changes easily by chance with different samples.

\textbf{Compensation and qualification}
Most papers do not reveal compensation for annotators and qualification filters, despite the fact that the current ethical standard recommends reporting basic statistics on time commitment and compensation \cite{pay_crowdworkers}. 

\textbf{Interface design}
The user interface design for annotation offers a high degree of freedom, and various choices can impact ratings. For instance, a constant image resolution of 256x256 was used in \cite{ding_cogview2_2022}, while a mixture of resolutions for different models were applied in \cite{inferring_2018_cvpr,stackgan_2017_iccv}. Moreover, a real image may be displayed side-by-side for reference as in \cite{ding_cogview2_2022,parti_2022_arxiv}.
Such details of the user interface are often undisclosed.
While researchers often share their model code, it is less common for human evaluation interfaces.
Designing instructions, tasks, and rating options is critical and requires substantial consideration.
The lack of reusable resources hinders the continuous improvement of human evaluation protocols and practices.

\section{Our design for text-to-image evaluation}
\label{sec:design}

We decided to use a crowdsourcing platform for human evaluation. The design of our evaluation task follows two principles.
Firstly, \textit{the task should be simple}.
We make it so simple that even inexperienced annotators can finish an instance of the task without extra effort to be familiar with it.
Secondly, \textit{evaluation results should be interpretable}.
Human evaluation is for helping researchers understand the models output; thus resulting annotation data should be useful for follow-up analysis. For example, the evaluation that produces relative rankings is difficult to interpret as each sample's rating depends on other samples. For better interpretability, direct scoring is preferred in recent human evaluations for natural language generation \cite{genie-2022,perils-2021,barrault-etal-2020-findings}.

\begin{figure*}[ht!]
    \centering
    \includegraphics[width=\linewidth,clip]{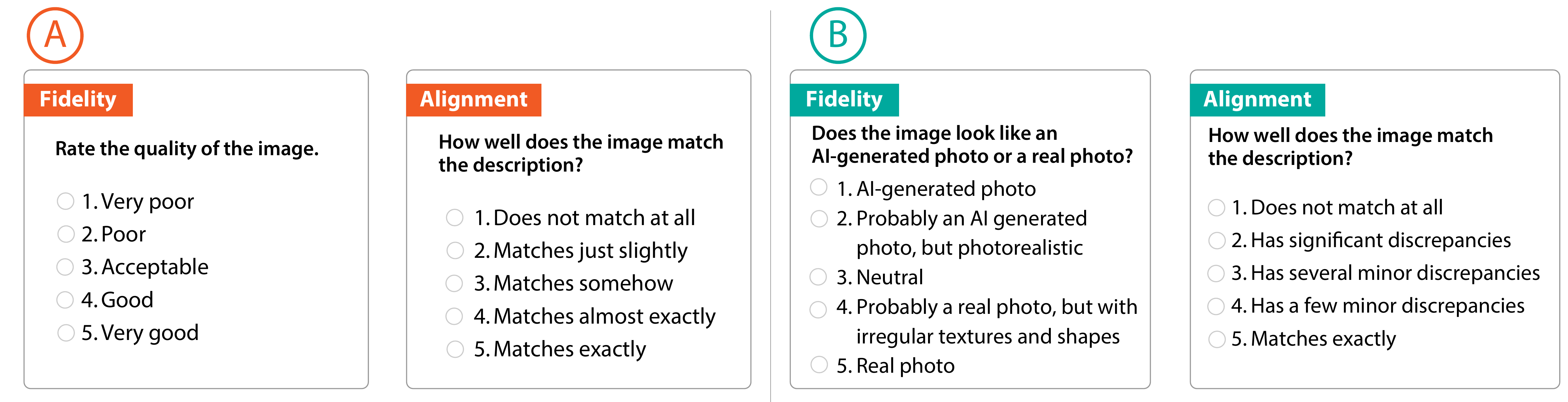}
\caption{Question and labels of two candidate task designs. A uses typical labels for a 5-point Likert scale. B's labels are more precise.}
\label{fig:qa_wording}
\end{figure*}

\subsection{Rating format}
There are two major options in rating methods: comparative and absolute.
Comparative evaluation, such as ranking generated images, is usually easier for annotators, and their ratings tend to be consistent.
However, comparative evaluation needs baseline models shared among all evaluation attempts.
At least currently, generative models enjoy rapid advancement, which can make baselines outdated in a short period. 
Another problem is its limited interpretability.
Comparative evaluation only tells relative goodness among a given set of images but does not care about the goodness among all. One image can win because either it is of high quality or the baseline is weak.
Lastly, comparative evaluation results in a relative ranking of models, and thus diachronic comparison (or comparison over time) among different evaluation results is almost infeasible.

Considering these limitations, we decided to adopt absolute evaluation as our basic rating method.
Yet absolute evaluation has some challenges.
Firstly, it is harder than comparative evaluation and tends to result in more noisy annotations.
Instructions, questions, and options (labels) must be carefully designed for quality control.

\subsection{Evaluation criteria and wording}
\label{sec:eval_criterion}

Our survey shows that many prior works employ \textit{fidelity} and \textit{alignment} as evaluation criteria. 
Fidelity means how well a generated image looks like a real photo, whereas alignment means how well a generated image matches the text.
We consider that these two criteria represent sufficiently the essential aspects of the quality of text-to-image generation and decide to follow the convention.

The wording of questions and option labels can largely affect annotators' labeling behavior.
A common indiscretion is to provide only endpoint labels, \eg, Likert scales 1: \textit{worst}, 5: \textit{best}, and the other options are unlabeled.

In our pilot data collection experiment, we tried two design candidates (\cref{fig:qa_wording}) to investigate the impact of concreteness in the questions and option labels.
(A) is a baseline task whose wording follows typical Likert scale labels.
(B) describes questions and options more specifically.
For alignment, (B) uses more detailed quantifiers than (A) to reduce subjectivity. For fidelity, (A) asks general quality and uses general labels, whereas (B) asks to judge if the image looks AI-generated or real.
A similar question was employed in \cite{hype-neurips2019} that asks binary judgment to distinguish real images from generated ones.

For the pilot data collection,
we sampled human annotations for 200 text-image pairs of COCO Captions \cite{ms-coco}. Among these 200 samples, 100 images are generated by Stable Diffusion \cite{stable_diffusion_Rombach_2022_CVPR} conditioned by the captions, while the other 100 images are real ones. 
The annotators get paid \$0.04 for each evaluation task.
We restrict the number of annotations per annotator to at most 40 to avoid a small set of annotators dominating all evaluation task instances.
We screen annotators with maturity, experience, and location qualification filters explained in \cref{sec:qual_ablation}.
As a result, Krippendorff's $\alpha$ for the alignment and fidelity questions in the case of (A) are 0.07 and 0.18, respectively, indicating high variations among annotators.
On the other hand, (B) achieved 0.39 for fidelity and 0.26 for alignment. Although not surprising, these results successfully confirm that more specific questions and labels lead to significantly better IAA. 
The following experiments use design (B).

\subsection{Qualifications}
\label{sec:qual_ablation}

Screening annotators is mandatory in AMT as
the annotator pool is diverse and global.
Some annotators do not have sufficient skills.
For example, language fluency is critical for our task since annotators are required to align English text and image.
% annotator's language skill also influences how they understand instructions.
AMT provides qualification filters that allow its users to employ annotators who meet task-specific needs. Meanwhile, it is not trivial to identify the sufficient and necessary set of qualification filters (too many filters may reduce the number of potential annotators). We thus experimentally explore qualification settings. 

We tested the following qualification filters.
\begin{enumerate}[label=\roman*)]
  \setlength{\parskip}{0cm}
  \setlength{\itemsep}{0cm}
    \item \textit{Maturity}: Over 18 years old and agreed to work with potentially offensive content.
    \item \textit{Experience}: Completed more than 5000 HITs with an approval rate larger than or equal to 99\%.
    \item \textit{Location}: Located in an English-speaking country\footnote{Following \cite{perils-2021}, we select annotators in US, Canada, UK, Australia, and New Zealand}.
    \item \textit{Skillfulness}: Passed a pre-task qualification test with three simple questions to confirm basic skills to assess image quality and to align text and image.\footnote{Examples of the qualification test are in the supplementary material.} 
    \item \textit{Master}: Good-performing and granted AMT Masters.
\end{enumerate}
As we are not fully aware of the content in generated images, we always use the maturity qualification. Although location qualification iii) is recommended in prior work in NLG \cite{perils-2021,genie-2022}, VPN can easily cheat this qualification and is a common practice among annotators \cite{shape-solution-2020}.
Also, it should be noted that a substantial group of residents in the US do not use English as their primary language.

\begin{table*}[t!]
\centering
\caption[Comparison of four combinations of qualification filters. Scores for fidelity and alignment are computed by first taking the mean over all three human ratings for each sample and then taking the mean over all samples. We compute Krippendorff's $\alpha$ as an IAA measure. Med.~time is the median time between successive submissions as a proxy for time to complete a single instance of the task.]{Comparison of four combinations of qualification filters. Scores for fidelity and alignment are computed by first taking the mean over all three human ratings for each sample and then taking the mean over all samples. We compute Krippendorff's $\alpha$ as an IAA measure\footnotemark. Med.~time is the median time between successive submissions as a proxy for time to complete a single instance of the task.}
\label{tab:qualification_ablation}
\begin{tabular}{@{}cccccccccccc@{}}
\toprule
\multicolumn{5}{c}{Qualification} & \multicolumn{3}{c}{Annotator performance}      & \multicolumn{2}{c}{Stable Diffusion} & \multicolumn{2}{c}{Real image} \\
\cmidrule(rl){1-5} \cmidrule(rl){6-8} \cmidrule(l){9-10} \cmidrule(l){11-12}
i & ii & iii& iv& v& Fidelity IAA & Alignment IAA & Med.~Time & Fidelity & Alignment & Fidelity & Alignment\\ \midrule
\checkmark&\checkmark& & & & 0.11 & 0.10      & 12.0      & 3.81      & 4.63       & 4.78     & 4.94      \\
\checkmark&\checkmark&\checkmark& & & 0.39     & 0.26      & 16.0      & 2.83      & 4.18       & 4.43    & 4.76     \\
\checkmark&\checkmark&\checkmark&\checkmark& & 0.37     & 0.40      & 20.0      & 2.71       & 4.23       & 4.32     & 4.67      \\
\checkmark& & & &\checkmark& 0.53     & 0.44      & 25.0      & 2.65       & 4.18       & 4.58    & 4.81     \\ \bottomrule
\end{tabular}
\end{table*}
\footnotetext{\highlight{A positive IAA value indicates that the ratings are more consistent than random annotations. For example, the coherence rating of NLG in \cite{perils-2021} achieves an IAA of 0.14.}}

To ablate the impact of qualification filters, we conduct pilot data collection whose configuration is the same as the one in \cref{sec:design}.
Three annotators gave ratings for each sample based on the questions and labels of (B) in \cref{sec:eval_criterion}.
For ablation, we published the task on the same day but with different combinations of qualification filters.

\Cref{tab:qualification_ablation} summarizes the results.
We observed that the annotator group with maturity and experience qualifications spent the shortest time per instance.
This may suggest inattention of the group, reflected in much lower IAA and much higher fidelity scores to generated images than the other groups.
Adding location and skillfulness qualification filters shows positive impacts in terms of IAA, and annotators take more time to complete tasks.
However, requiring more qualifications, especially skillfulness qualification, overly downsizes the annotator pool and prolongs the time to collect all annotations.
AMT Masters group by itself achieved a relatively high IAA and did not degrade the total annotation time.
Therefore, we use the master qualification filter in the following experiments.

\section{Human evaluation of existing models}

\begin{table*}[t!]
\begin{tabular}{lr}
\begin{minipage}[t]{0.6\linewidth}
% \centering
\caption{Human and automatic evaluation of generated and real images on MS-COCO. Rankings by human evaluation and automatic evaluation are misaligned.}
\label{tab:coco-leaderboard}
\begin{tabular}{@{}rcccc@{}}
\toprule
\multirow{2}{*}{model} & \multicolumn{2}{c}{Human}        & \multicolumn{2}{c}{Automatic}    \\
                       & Fidelity $\uparrow$             & Alignment $\uparrow$           & FID $\downarrow$            & CLIPScore $\uparrow$          \\ \midrule
LAFITE \cite{lafite_zhou2022}                 & 1.77                     &  3.73                    & 34.46           & 0.82                \\
GLIDE \cite{glide_icml_2022}                  & 2.56                     &  2.96                    & 39.80           & 0.68                \\
CogView2 \cite{ding_cogview2_2022}               & 2.19                     &  3.55                    & 29.57           & 0.68                \\
Stable Diffusion \cite{stable_diffusion_Rombach_2022_CVPR}       &  3.09                    &  4.35                    & 32.19           & 0.78                \\ \midrule
Real Image             & 4.49                & 4.78                  &      ---       & 0.76                \\ \bottomrule
\end{tabular}
\end{minipage}
\hspace{0.1em}
\begin{minipage}[t]{0.36\linewidth}
\caption{Human evaluation results on DrawBench\cite{imagen_drawbench} and PartiPrompts \cite{parti_2022_arxiv}.}
\label{tab:drawbench_parti}
\begin{tabular}{@{}rcc@{}}
    \toprule
           & Fidelity & Alignment \\ \midrule
\multicolumn{3}{l}{\small \textsc{DrawBench}} \\
    LAFITE \cite{lafite_zhou2022} &   1.82       &   3.35        \\
    GLIDE \cite{glide_icml_2022} &   2.67       &   3.41        \\
    CogView2 \cite{ding_cogview2_2022}&   2.27       &   3.29        \\
    Stable Diffusion \cite{stable_diffusion_Rombach_2022_CVPR} &  2.87        &   3.99        \\ \midrule
\multicolumn{3}{l}{\small \textsc{PartiPrompt}} \\
LAFITE \cite{lafite_zhou2022}& 1.93     & 3.48      \\
GLIDE \cite{glide_icml_2022}& 2.71     & 3.38      \\
CogView2 \cite{ding_cogview2_2022}& 2.34     & 3.49      \\
Stable Diffusion \cite{stable_diffusion_Rombach_2022_CVPR}& 3.05     & 4.07      \\ \bottomrule
\end{tabular}
\end{minipage}
\end{tabular}
\end{table*}

We evaluate four text-to-image generative models in the zero-shot setting. We do not use prompt engineering or sampling for screening generated images.

\begin{description}
  \setlength{\parskip}{0cm}
  \setlength{\itemsep}{0cm}
\item[Lafite \cite{lafite_zhou2022}]  is a GAN model with CLIP text encoding. A model trained on Google CC3M \cite{cc3m_google} is used\footnote{https://github.com/drboog/Lafite}.
\item[GLIDE \cite{glide_icml_2022}] is a text-to-image diffusion model. We used the publicly-available lightweight variant\footnote{https://github.com/openai/glide-text2im}. We follow the authors' parameter settings.
\item[CogView2 \cite{ding_cogview2_2022}] is a hierarchical transformer-based model trained on a large-scale corpus of images with Chinese and English captions\footnote{https://github.com/THUDM/CogView2}.
\item[Stable Diffusion \cite{stable_diffusion_Rombach_2022_CVPR}] is another diffusion model. We use the v1.4 model hosted by Hugging Face\footnote{https://huggingface.co/CompVis/stable-diffusion-v1-4}.
\end{description}

%\textbf{Dataset}
We use as sources of captions and images the datasets that are widely used in the literature; COCO Captions \cite{ms-coco}, DrawBench \cite{imagen_drawbench} and PartiPrompts \cite{parti_2022_arxiv}.
COCO Captions provides images and five annotated captions for each image. We randomly pick one out of five as a prompt to generate an image. We discard invalid captions, such as ``unable to see this image in this particular hit.''
DrawBench and PartiPrompts are prompt datasets tailored for benchmarking text-to-image generative models.

%\textbf{Crowdsourcing setups}
We follow the setting in \cref{sec:design}, although annotators get \$0.05 for each instance of the task and the limitation of annotations per annotator was increased to 250. 
%We monitor a time difference between successive submissions as an approximation of time to complete one task.

\highlight{
We summarize human ratings by first taking the mean over all human ratings for each sample and then averaging over all samples. Our fidelity evaluation is interpreted as a Mean Opinion Score (MOS) test \cite{itu-t-p800} for visual quality, whereas alignment one is another test but similar to MOS.
}

\begin{figure*}[t!]
    \centering
    \includegraphics[width=0.9\linewidth,clip]{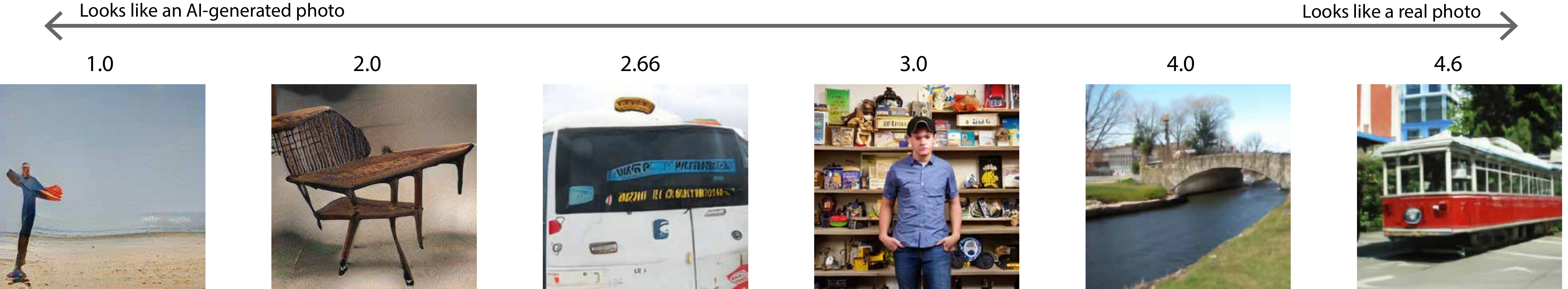}\\
    \includegraphics[width=0.9\linewidth,clip]{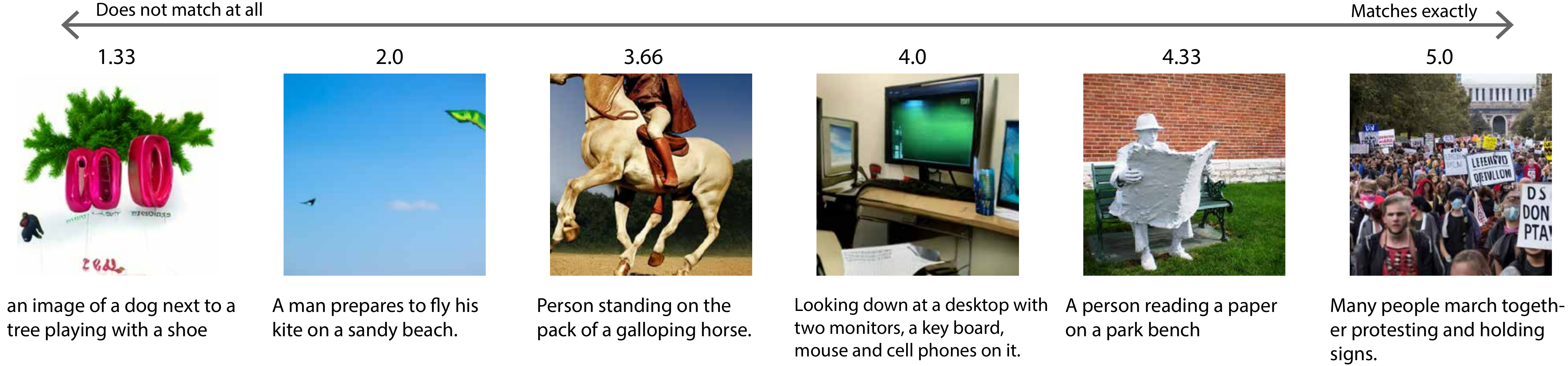}
    \caption{Generated or real images and their human ratings of fidelity and alignment. The scores are the mean of three annotators' ratings.}
    \label{fig:coco-examples}
\end{figure*}

\subsection{Evaluation results}
On COCO Captions, we collected annotations for images generated by the four models for 1,000 captions and real images of COCO Captions, resulting in 15,000 annotations.
Krippendorff's $\alpha$ of the fidelity and alignment ratings are 0.41 and 0.48, respectively.
148 annotators participated in total, and the average number of tasks per annotator was 101.4.
The median time spent on one task is 18 seconds; that is, the expected hourly wage is \$10, which is compatible with an ethical recommendation \cite{pay_crowdworkers}.
Collecting 15,000 annotations took 30 hours.
Examples of annotated caption-image pairs are shown in \cref{fig:coco-examples}.

Human evaluation results on COCO Captions are summarized in \cref{tab:coco-leaderboard}.
Real images in this dataset are preferred by human annotators in terms of both fidelity and alignment, and Stable Diffusion is the second best.

DrawBench provides 200 prompts. We generated images for all, resulting in 2,400 annotations.
Krippendorff's $\alpha$ for fidelity and alignment are 0.13 and 0.19.
The drop of IAA compared with COCO Captions can be due to the increase of difficulty in evaluation.
DrawBench is a collection of challenging textual prompts including long text, rare words, misspellings \etc.
Models often fail for such prompts, and annotators experience difficulties in evaluating significant failures and complex text.
40 annotators participated in total, and the average number of instances per annotator is 60.
The median time spent on a single instance is 14 seconds, so the expected hourly wage is \$12.9.
The overall time required to complete annotations was 1.7 hours.

On PartiPrompts, we collected annotations for images generated by the four models for 1,337 captions, resulting in 16,044 annotations. Krippendorff's $\alpha$ of the fidelity and alignment ratings are 0.21 and 0.40, respectively. 181 annotators participated in total, and the average number of tasks per annotator was 87.2. The median time spent on one task is 18 seconds. Collecting annotations took 48 hours.

The results on DrawBench and PartiPrompt are in \cref{tab:drawbench_parti}. The model ranking in terms of fidelity is the same as COCO Captions. Stable Diffusion significantly outperform other models on both DrawBench and PartiPrompt in terms of alignment. On the other hand, other models did not show statistically significant difference. More detailed results are in the supplementary material.

\subsection{Agreement between automatic measures and human evaluation}
\begin{figure}[t!]
    \centering
    \includegraphics[width=\linewidth,clip]{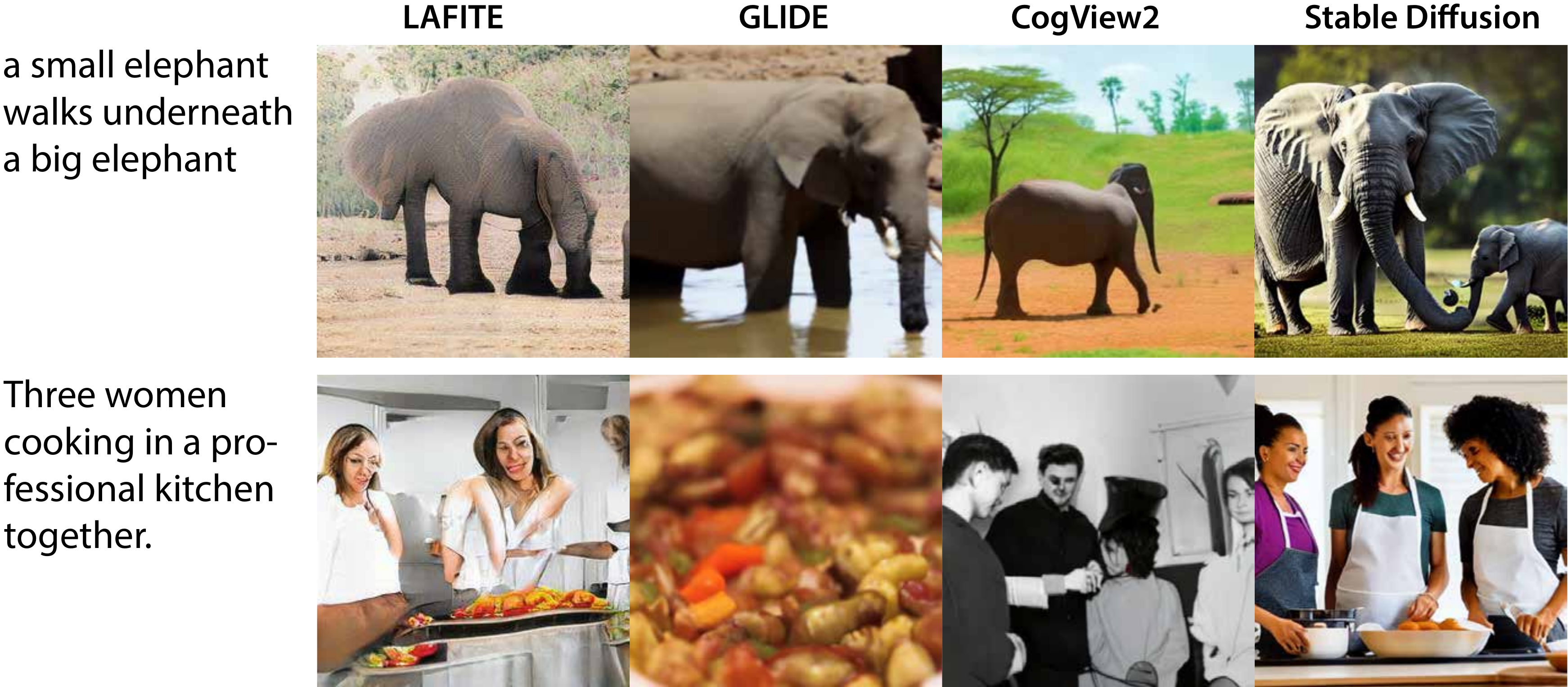}
    \caption{Examples of input captions and generated images.}
    \label{fig:generated_image}
\end{figure}
\paragraph{Fr\'echet Inception Distance}

\Cref{tab:coco-leaderboard} shows FID values. 
Many prior works reported that FID does not align with perceived quality \cite{ding_cogview2_2022,cafd2018,aliased_resizing_2022}, and our human evaluation supports their claim.
FID ranks CogView2 the best, while human annotators rated Stable Diffusion the best among the models in terms of fidelity and CogView2 in the third place.
\Cref{fig:generated_image} shows generated samples by the models.
We observe that CogView2 produces more artifacts than Stable Diffusion (\cref{fig:generated_image}).
FID fails to capture such irregular textures.
The ranks of LAFITE and GLIDE are also inconsistent between FID and human evaluation.
These results suggest that validation of fidelity solely relying on FID can lead to inconsistency with human perception.

\paragraph{CLIPScore}

\begin{figure}[t!]
    \centering
    \includegraphics[width=0.9\linewidth,clip]{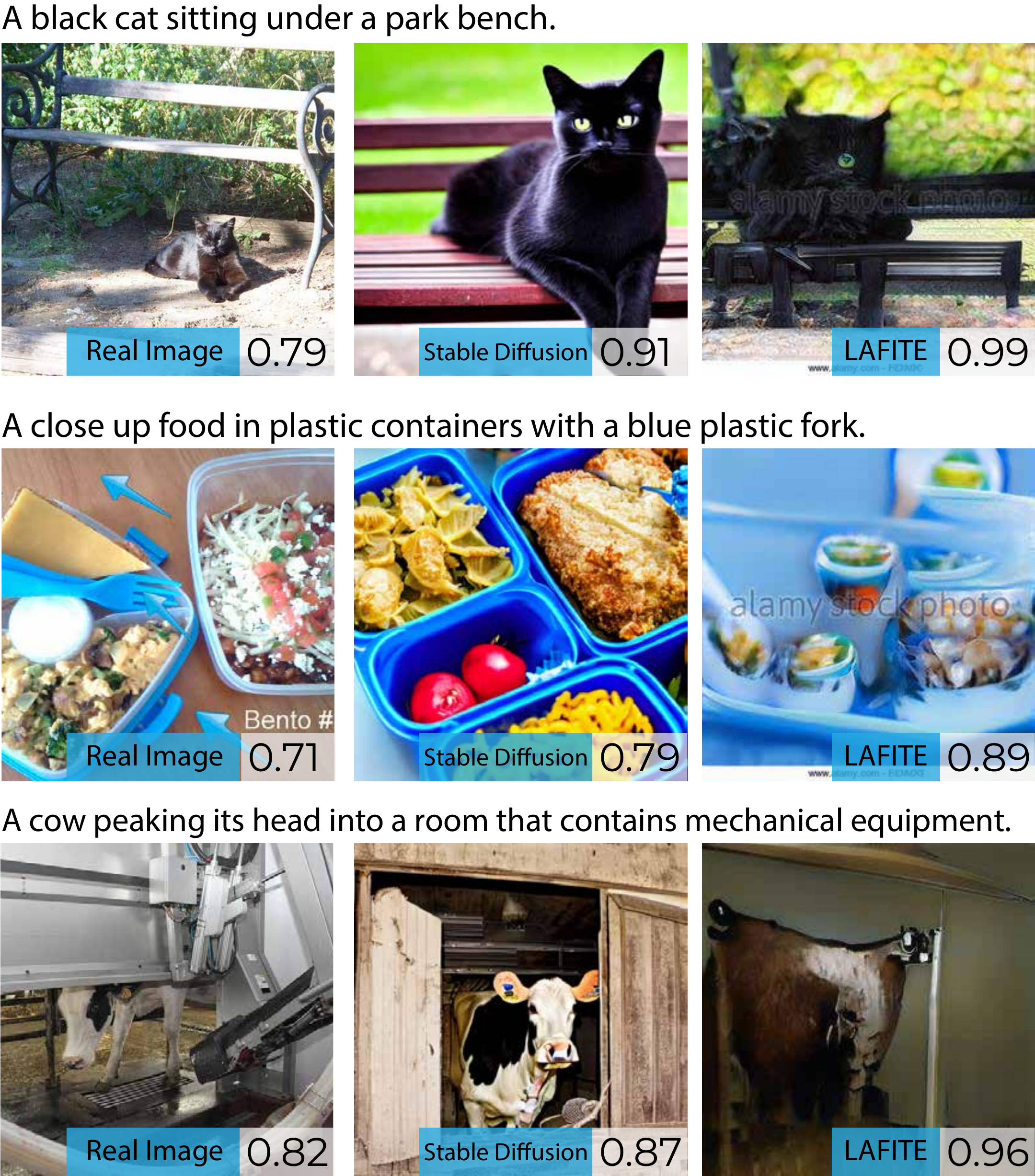}
    \caption{Below each caption, a real image and two generated images using the caption are displayed. Their sample-level CLIPScore is displayed in the bottom right of each. LAFITE achieves high scores, which are counter-intuitive.}
    \label{fig:clipscore_example}
\end{figure}

We investigate CLIPScore \cite{hessel-etal-2021-clipscore}, a recently proposed automatic measure for text-to-image alignment.
Interestingly, LAFITE and Stable Diffusion achieve higher scores than image and caption pairs from COCO Captions, while human annotators rate the real images the best in terms of alignment as shown in \cref{tab:coco-leaderboard}.
However, as observed in \Cref{fig:clipscore_example}, better CLIPScore does not necessarily mean better alignment, and the scale of the scores does not represent how well the image and text are aligned.

LAFITE's score is substantially higher than others.
This may be because LAFITE involves optimization with respect to the CLIP-based GAN loss as discussed in \cite{glide_icml_2022}.
% The result suggest that CLIPScore can be gameable when the measure is targetd for optimization.
On the other hand, Stable Diffusion gives higher scores without optimization in the CLIP space.
However, CLIPScore does not discriminate the alignment performance between real images and Stable Diffusion generated images, while human annotators rate real images higher.

The comparison to human evaluation and qualitative result clarifies the limitation of CLIPScore.
Evaluation measures are often the objectives in optimization \cite{Anderson2017up-down}. Such optimization, however, suffers from adversarial effects.
LAFITE, which optimizes CLIPScore, seems to exhibit this problem.
%Evaluation measures are often targeted for optimization, and as a result, the measure can become unable to reflect real progress.
Another problem is that CLIPScore cannot discriminate real images from images generated by Stable Diffusion, despite human annotators being able to distinguish them.
This suggests that even with authentic caption-image pairs, there may be a gap in the CLIP space, which is not surprising, and images generated by Stable Diffusion are already within this margin.
That is, CLIPScore is already saturated and may no longer be useful to evaluate state-of-the-art generative models. 

\subsection{Effect of sample size}
\label{sec:sample_size}
%This section investigates the design of reliable human-annotated datasets.
The sample size is a major factor in experimental design. That is, the reliability of conclusions depends much on the number of human ratings, while it is directly reflected in monetary costs.
The sample size here involves (1) the number of prompts and (2) the number of annotators who evaluate the same image.

\begin{figure}[t!]
    \centering
    \includegraphics[width=\linewidth,clip]{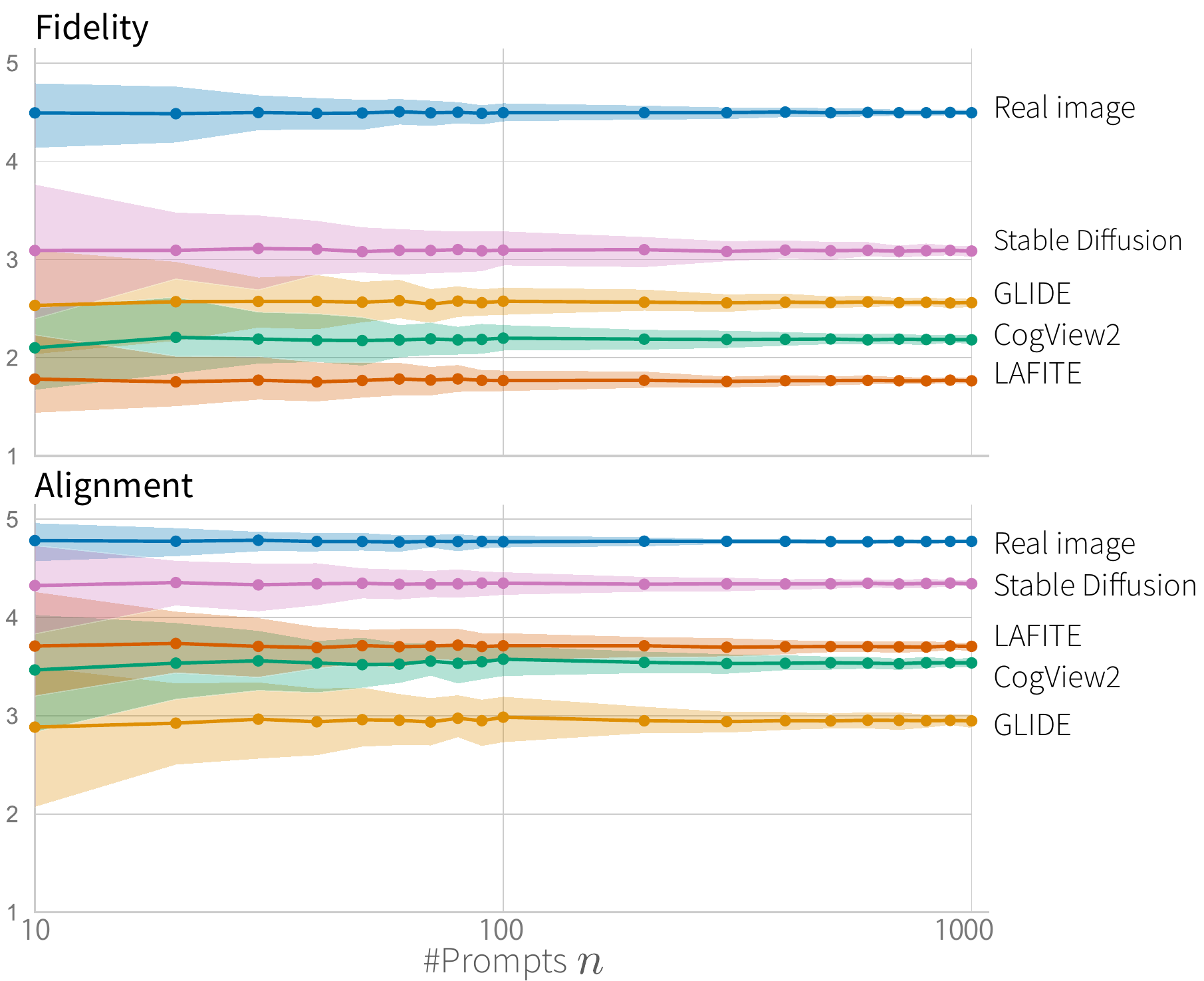}
    \caption{Effect of the number of prompts in the evaluation dataset. We compute the fidelity and alignment scores over sampled $n$ prompts. Each data point represents the mean over the 500 trials, and the colored area represents the 5\%-95\% percentile interval. With a small number of test prompts, human evaluation can produce different rankings of the models by chance.}
    \label{fig:effect_of_number_of_prompts}
\end{figure}

\textbf{Number of prompts}
We repeatedly computed the fidelity and alignment scores 500 times over $n$ samples randomly drawn from 1000 COCO Captions.
\Cref{fig:effect_of_number_of_prompts} shows the mean over the 500 trials with the 5\%-95\% percentile interval.
For both fidelity and alignment scores, there are large overlaps of the intervals when $n$ is small.
This suggests that conclusions drawn from small $n$ may be unstable and easily flipped depending on the choice of prompts.
With our choice of four models, we need more than 100 prompts to obtain consistent conclusions. % with that using the entire prompts---this implies the unreliability of the experiments with less than 100 prompts in the previous studies. 
\Cref{fig:effect_of_number_of_prompts} gives a useful insight into the relationship between the difference in scores and the conclusive number of prompts.

\begin{figure}[t!]
    \centering
    \includegraphics[width=\linewidth,clip]{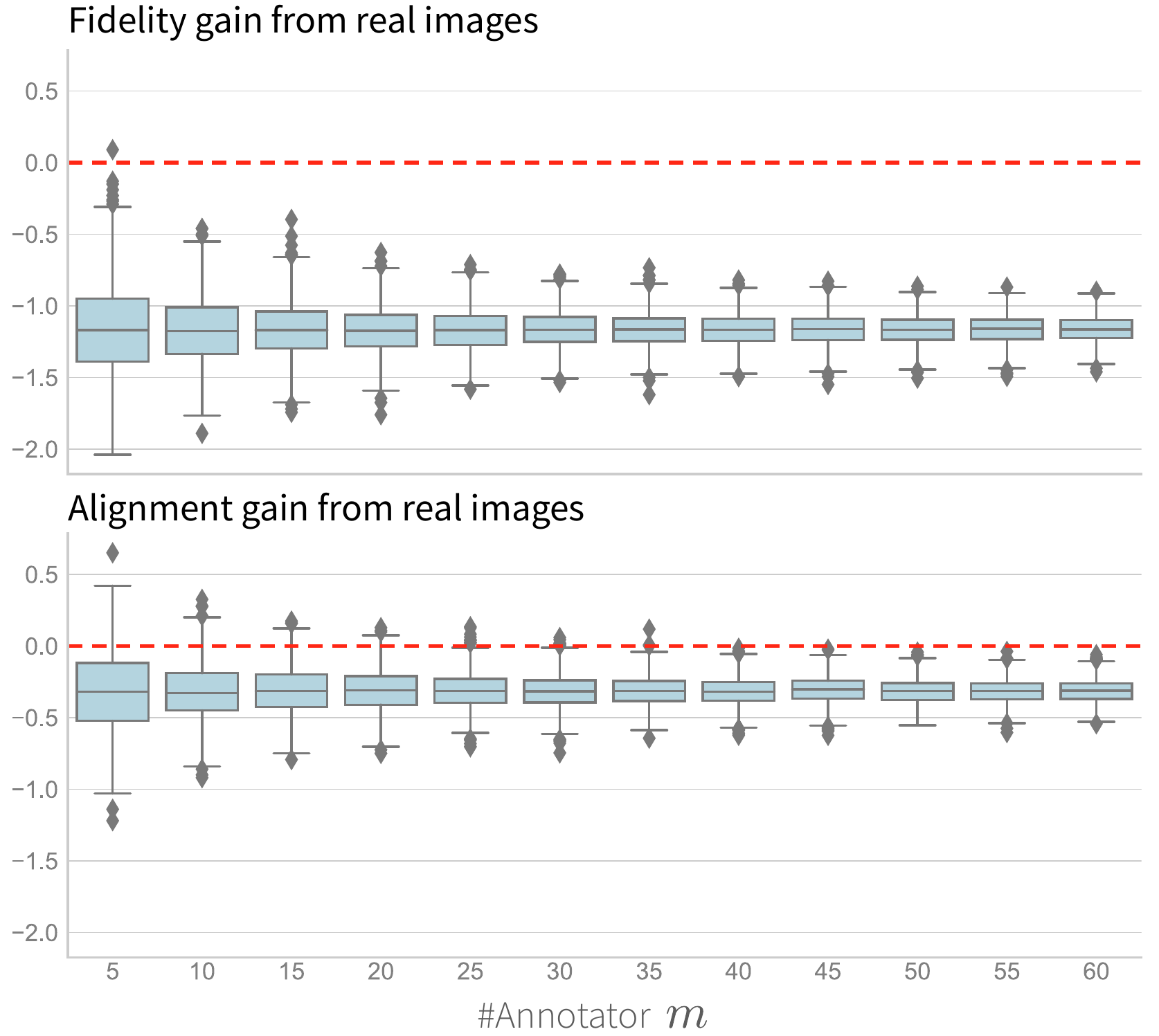}
    \caption{Effect of the number of raters per sample. The gain below the red dashed line indicates that real images outperform Stable Diffusion. The positive gains with few ratings demonstrate that few ratings per sample lead to instability.}
    \label{fig:effect_of_number_of_raters}
\end{figure}

\textbf{Number of annotators for a single image}
We selected 13 captions from COCO Captions as prompts\footnote{The selection strategy is described in the supplementary material.} and collected 60 human ratings for each of the corresponding 13 real images and images generated by Stable Diffusion. 
%We also investigate the effect of the number of raters $m$ for a single text-image pair.
%To carry out this experiment, we additionally collected another subset of human-annotated data, in which different 60 raters judged an image obtained from the real image baseline and Stable Diffusion for 13 selected prompts\footnote{The detail of the selection strategy for the prompts is described in the appendix.}.
We randomly sampled $m$ ratings out of 60 and computed the scores 500 times. This time, we computed the gain from the real images and Stable Diffusion images for each trial\footnote{The gains are mostly negative as real images' scores are mostly better.}. 
\Cref{fig:effect_of_number_of_raters} shows their mean and 5\%-95\% percentile interval for $m$.  %\in \{5, 10, \dots, 60\}$ 
%in the same procedure as for $n$.
%The x- and y-axes in the figures indicate the number of raters and the performance gain of Stable Diffusion from the real image baseline, respectively, in terms of the fidelity (top) and alignment (bottom); note here that the gain from the baseline should be negative.
There are similar trends in both fidelity and alignment: Evaluation with fewer ratings often leads to instability.
Particularly for alignment, we can occasionally draw a conclusion of the superiority of Stable Diffusion to real images.
In the scenario of the lack of per-sample annotations, we may avoid unreliable conclusions by reporting (1) the statistical significance and (2) the effect size (\eg, Hedge's $g$~\cite{hedges1981distribution}).
\highlight{These reliability checks are recommended especially for low-budget experiments.}
Taking into account the deviation of the scores, ensuring a gain greater than 0.5 may be an easy way to conservative conclusions.

\section{Discussions}
\subsection{Recommendations for crowdsourcing}

The comparison between automatic measures and human evaluation reveals that current automatic measures are insufficient to represent human perception, and relying solely on them risks the reliability of conclusions.
A careful discussion is recommended based on both automatic measures and human evaluation, of which protocol is not yet matured.
We thus propose a guideline for better evaluation.

\textbf{Reporting experimental details for transparency}
Due to the difficulty of controlling annotation quality, results can vary over different runs of the same annotation process \cite{perils-2021}. Literature should provide detailed description of the annotation for verifiability and reproducibility.
Based on the recommendations for reporting experiments using crowdsourcing \cite{mturk_research}, we offer a sample template for reporting human evaluation settings in the supplementary material.

\textbf{Understanding crowdworkers}
Monitoring annotators' behavior is an essential way to quality control.
Using automation tools for efficiently completing many tasks is a common practice for crowdworkers \cite{Kaplan_Saito_Hara_Bigham_2018}, but these tools cause unnatural submission logs.
For example, we observed many submissions that supposedly come from automating task approval.
There are many other tools, \eg, for annotation interface optimization, time management, and displaying requesters' reputations.
Tasks that are incompatible with such tools limit annotator pools.

% limitation
\subsection{Limitations and future work}
% future work
Our human evaluation protocol considers fidelity and alignment as two important criteria. However, there are other important criteria to consider, such as undesired bias in generated images, which is crucial in various applications. It is also important to evaluate this aspect.

We focus on natural images, but text-to-image generation can also create artwork. Different domains may require different evaluation criteria, \eg, likability, and aesthetics.

\highlight{AMT master qualification has limitations: i) it is limited to AMT, and ii) criteria of master qualification is unclear. Combining other qualifications substitutes for AMT master qualification to some extent. Post-hoc filtering of annotators based on IAA will also improve annotation quality.}

\highlight{A critical challenge in human evaluation is that collecting annotations large enough to get a reliable result is still expensive. One plausible remedy is to involve sampling techniques to effectively estimate models' performance with a smaller number of samples.}

\highlight{

}

\section{Conclusion}

Our survey on human evaluations in recent text-to-image literature reveals reliability and transparency issues in human evaluation. 
We thus develop a human evaluation protocol for text-to-image generation. %State-of-the-art models are evaluated with our evaluation protocol using crowdsourcing and automatic measures. 
Our experiments demonstrated that automated measures do not align with human perception and are already getting outdated. The community needs to keep updating the automatic measures to catch up with the evolution of generative models. 
Yet, human evaluation itself is a challenging problem, and our design may not be optimal.
We share our code with the community for continuous improvement in efficiency and coverage.

\paragraph{Acknowledgement}
This work is partly supported by JST CREST Grant No.\@ JPMJCR20D3, JST FOREST Grant No.\@ JPMJFR216O, JSPS KAKENHI Grant-in-Aid for Scientific Research (A), the Academy of Finland projects 324346 and 353139.

%%%%%%%% REFERENCES
{\small
\bibliographystyle{ieee_fullname}
\bibliography{egbib}
}

\end{document}

\title{Supplementary Material}
\renewcommand\thefigure{\Alph{figure}}
\renewcommand\thetable{\Alph{table}}
\author{
}
\maketitle
%%%%%%%%% BODY TEXT
\section{Details of Reviewed Papers}
We count the number of papers in which a certain experimental detail is reported. \cref{tab:survey_paper_stats} shows that many papers fail to describe important details including the number of annotations and the number of ratings per sample. For annotation quality assessment, no paper report inter-annotator-agreement. The number of papers that employ certain types of evaluation criteria and rating method are also summarized in \cref{tab:survey_paper_stats}. We find that evaluation criteria and how to collect ratings vary from one paper to another.
The full list of surveyed papers is in \cref{tab:paper_list}.

\begin{table}[H]
\caption{The number of papers that report the details. Critical details are often omitted. The way of rating varies by paper.}
\label{tab:survey_paper_stats}
\centering
\begin{tabular}{@{}llc@{}}
\toprule
                     &                   & Count \\ \midrule
Numbers              & \# samples        &  18                 \\
                     & \# raters         &  11                 \\
                     & \# rates / sample &  4                  \\ 
Task Design          & Question          &  20                 \\
                     & Label             &  20                 \\
                     & Instruction       &  5                  \\
Quality check        & IAA               &  0                  \\
Annotator pool       & Crowdsourcing     &  3                  \\
                     & NA                &  17                 \\ 
\makecell{Crowdsourcing\\parameters}    & qualifications    &  2                  \\
                     & compensations     &  3                  \\
Criteria        & Quality              &  18                  \\
                & Relevance to prompts &  14                  \\
                & Others               &  2                   \\
Types of rating & Choice (w/wo ties)   &  10                  \\
                & Ranking              &  9                   \\
                & Numeric              &  3                   \\ \bottomrule
\end{tabular}
\end{table}

\begin{table}[H]
\caption{Full list of surveyed papers.}
\label{tab:paper_list}
\begin{tabular}{@{}lcc@{}}
\toprule
Title & Year & Venue \\ \midrule
An Image is Worth One Word: Personalizing Text-to-Image Generation using Textual Inversion \cite{gal_image_2022} & 2022 & ArXiv \\
AttnGAN: Fine-Grained Text to Image Generation With Attentional Generative Adversarial Networks \cite{xu_attngan_2018}& 2018 & CVPR \\
CogView: Mastering Text-to-Image Generation via Transformers \cite{ding_cogview_2021}& 2021 & NeurIPS \\
CogView2: Faster and Better Text-to-Image Generation via Hierarchical Transformers \cite{ding_cogview2_2022}& 2022 & ArXiv \\
Controllable Text-to-Image Generation \cite{li_controllable_2019}& 2019 & NeurIPS \\
CookGAN: Causality Based Text-to-Image Synthesis \cite{zhu_cookgan_2020}& 2020 & CVPR \\
CPGAN: Content-Parsing Generative Adversarial Networks for Text-to-Image Synthesis \cite{liang_cpgan_2020}& 2020 & ECCV \\
Cross-Modal Contrastive Learning for Text-to-Image Generation \cite{zhang_cross-modal_2021}& 2021 & CVPR \\
DAE-GAN: Dynamic Aspect-Aware GAN for Text-to-Image Synthesis \cite{ruan_dae-gan_2021}& 2021 & ICCV \\
DF-GAN: A Simple and Effective Baseline for Text-to-Image Synthesis \cite{tao_df-gan_2022}& 2022 & CVPR \\
DM-GAN: Dynamic Memory Generative Adversarial Networks for Text-To-Image Synthesis \cite{zhu_dm-gan_2019}& 2019 & CVPR \\
DreamBooth: Fine Tuning Text-to-Image Diffusion Models for Subject-Driven Generation \cite{ruiz_dreambooth_2022}& 2022 & ArXiv \\
Dual Adversarial Inference for Text-to-Image Synthesis \cite{lao_dual_2019}& 2019 & ICCV \\
GLIDE: Towards Photorealistic Image Generation and Editing with Text-Guided Diffusion Models \cite{glide_icml_2022}& 2022 & ArXiv \\
High-Resolution Image Synthesis With Latent Diffusion Models \cite{rombach_high-resolution_2022}& 2022 & CVPR \\
Imagic: Text-Based Real Image Editing with Diffusion Models \cite{kawar_imagic_2022}& 2022 & ArXiv \\
Inferring Semantic Layout for Hierarchical Text-to-Image Synthesis \cite{inferring_2018_cvpr}& 2018 & CVPR \\
Make-A-Scene: Scene-Based Text-to-Image Generation with Human Priors \cite{gafni_make--scene_2022}& 2022 & ECCV \\
MirrorGAN: Learning Text-To-Image Generation by Redescription \cite{qiao_mirrorgan_2019}& 2019 & CVPR \\
Object-Driven Text-To-Image Synthesis via Adversarial Training \cite{li_object-driven_2019}& 2019 & CVPR \\
Photographic Text-to-Image Synthesis With a Hierarchically-Nested Adversarial Network \cite{zhang_photographic_2018}& 2018 & CVPR \\
Photorealistic Text-to-Image Diffusion Models with Deep Language Understanding \cite{saharia_photorealistic_2022}& 2022 & ArXiv \\
RiFeGAN: Rich Feature Generation for Text-to-Image Synthesis From Prior Knowledge \cite{cheng_rifegan_2020}& 2020 & CVPR \\
Scaling Autoregressive Models for Content-Rich Text-to-Image Generation \cite{yu_scaling_2022}& 2022 & ArXiv \\
Semantics Disentangling for Text-To-Image Generation \cite{yin_semantics_2019}& 2019 & CVPR \\
Semantics-Enhanced Adversarial Nets for Text-to-Image Synthesis \cite{tan_semantics-enhanced_2019}& 2019 & ICCV \\
StackGAN: Text to Photo-Realistic Image Synthesis with Stacked Generative Adversarial Networks \cite{zhang_stackgan_2017}& 2017 & ICCV \\
StackGAN++: Realistic Image Synthesis with Stacked Generative Adversarial Networks \cite{zhang_stackgan_2018}& 2018 & ArXiv \\
StoryDALL-E: Adapting Pretrained Text-to-Image Transformers for Story Continuation \cite{maharana_storydall-e_2022}& 2022 & ECCV \\
StyleT2I: Toward Compositional and High-Fidelity Text-to-Image Synthesis \cite{li_stylet2i_2022}& 2022 & CVPR \\
Text to Image Generation With Semantic-Spatial Aware GAN \cite{liao_text_2022}& 2022 & CVPR \\
Text-to-Image Synthesis Based on Object-Guided Joint-Decoding Transformer \cite{wu_text--image_2022}& 2022 & CVPR \\
TISE: Bag of Metrics for Text-to-Image Synthesis Evaluation \cite{avidan_tise_2022}& 2022 & ECCV \\
Towards Language-Free Training for Text-to-Image Generation \cite{zhou_towards_2022}& 2022 & CVPR \\
Trace Controlled Text to Image Generation \cite{yan_trace_2022}& 2022 & ECCV \\
Vector Quantized Diffusion Model for Text-to-Image Synthesis \cite{gu_vector_2022}& 2022 & CVPR \\
Zero-Shot Text-to-Image Generation \cite{ramesh_zero-shot_2021}& 2021 & PMLR \\

\bottomrule
\end{tabular}
\end{table}

\newpage
\section{Annotation interface}
\begin{figure}[H]
    \centering
    \includegraphics[width=\linewidth,clip]{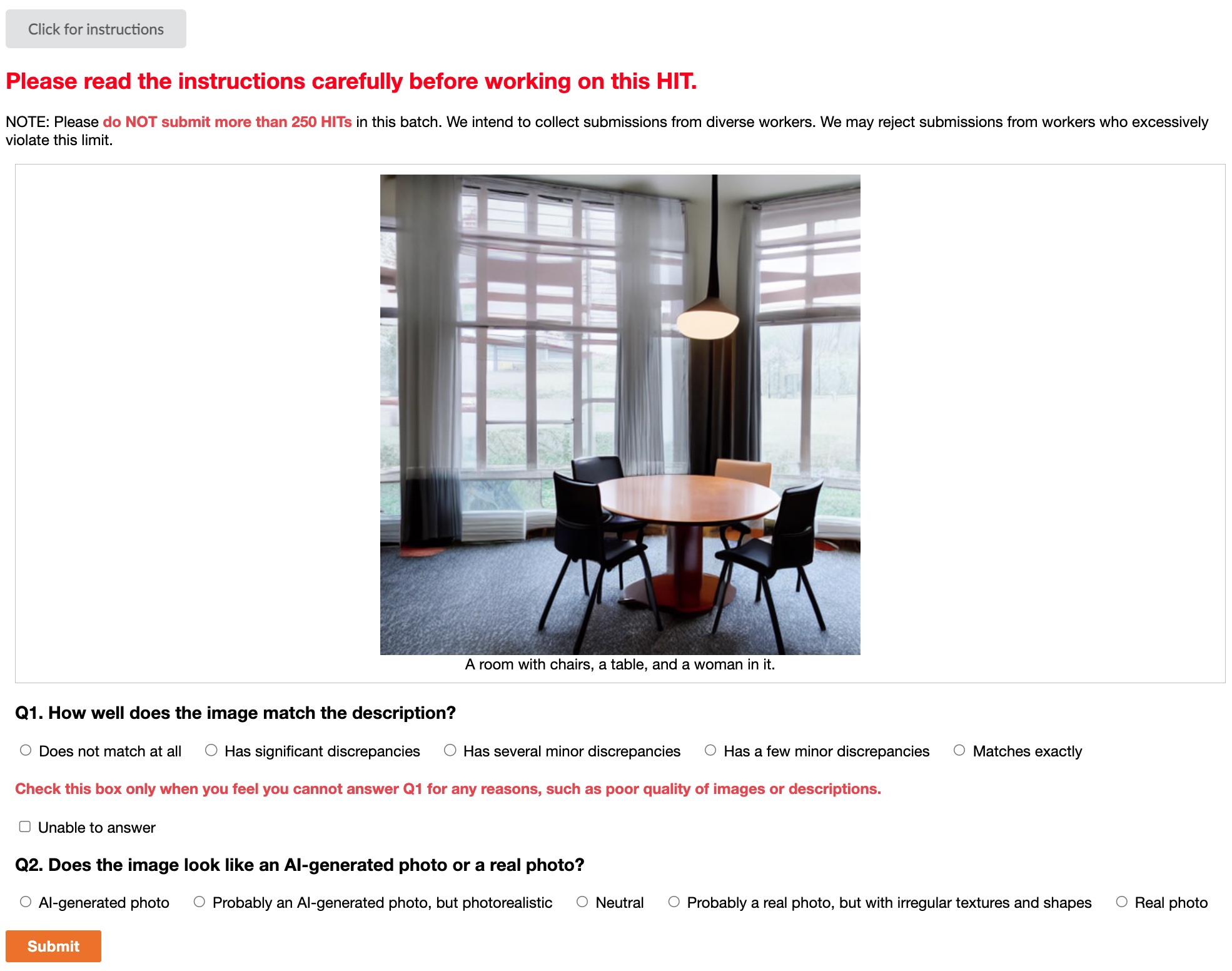}
    \caption{Annotation interface. Annotators rate the image in terms of fidelity and alignment.}
    \label{fig:anno_interface}
\end{figure}

\begin{figure}[H]
    \centering
    \includegraphics[width=0.8\linewidth,clip]{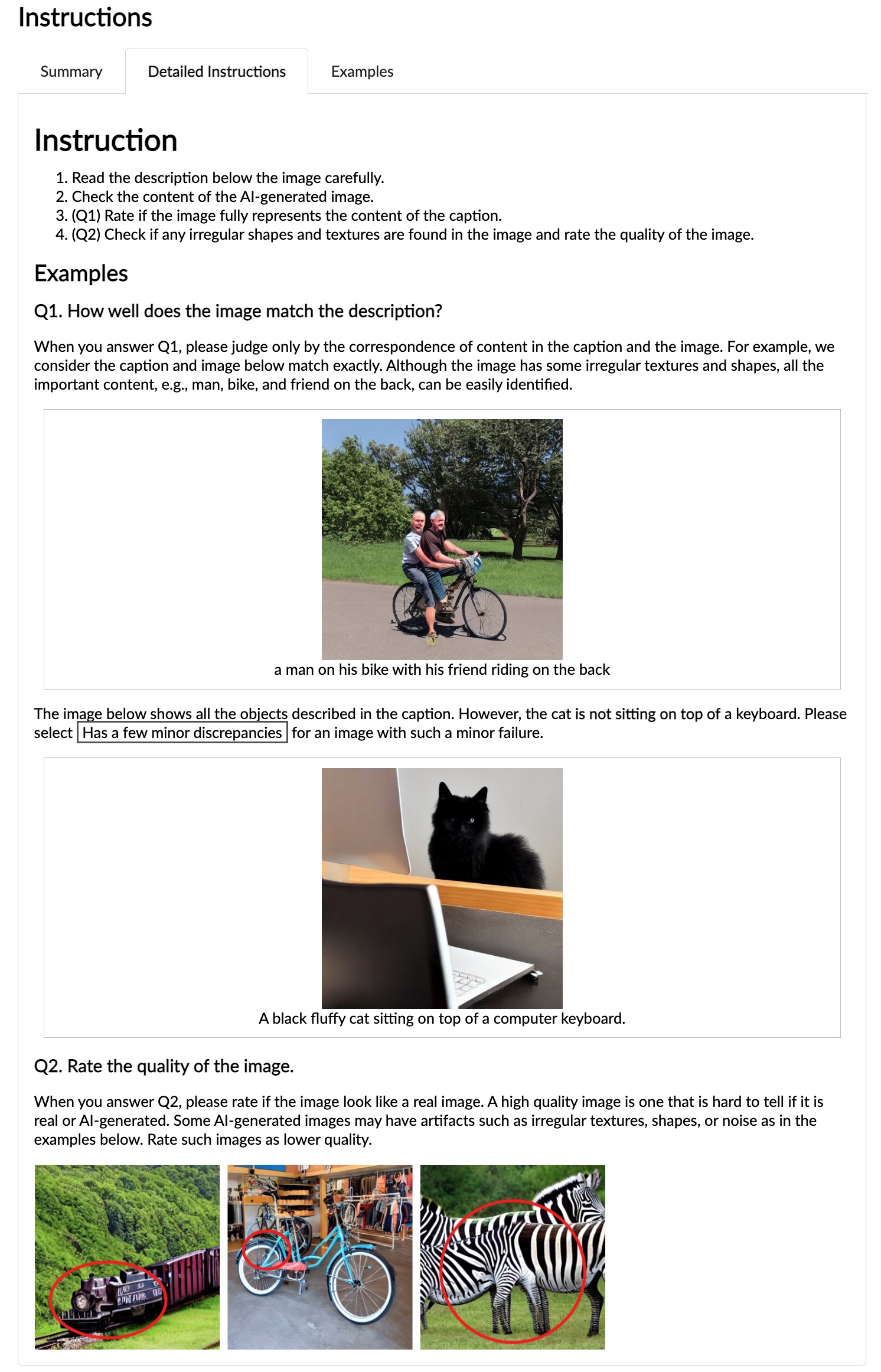}
    \caption{Instructions provided to the annotators.}
    \label{fig:instruction}
\end{figure}

\begin{figure}[H]
    \centering
    \includegraphics[width=0.8\linewidth,clip]{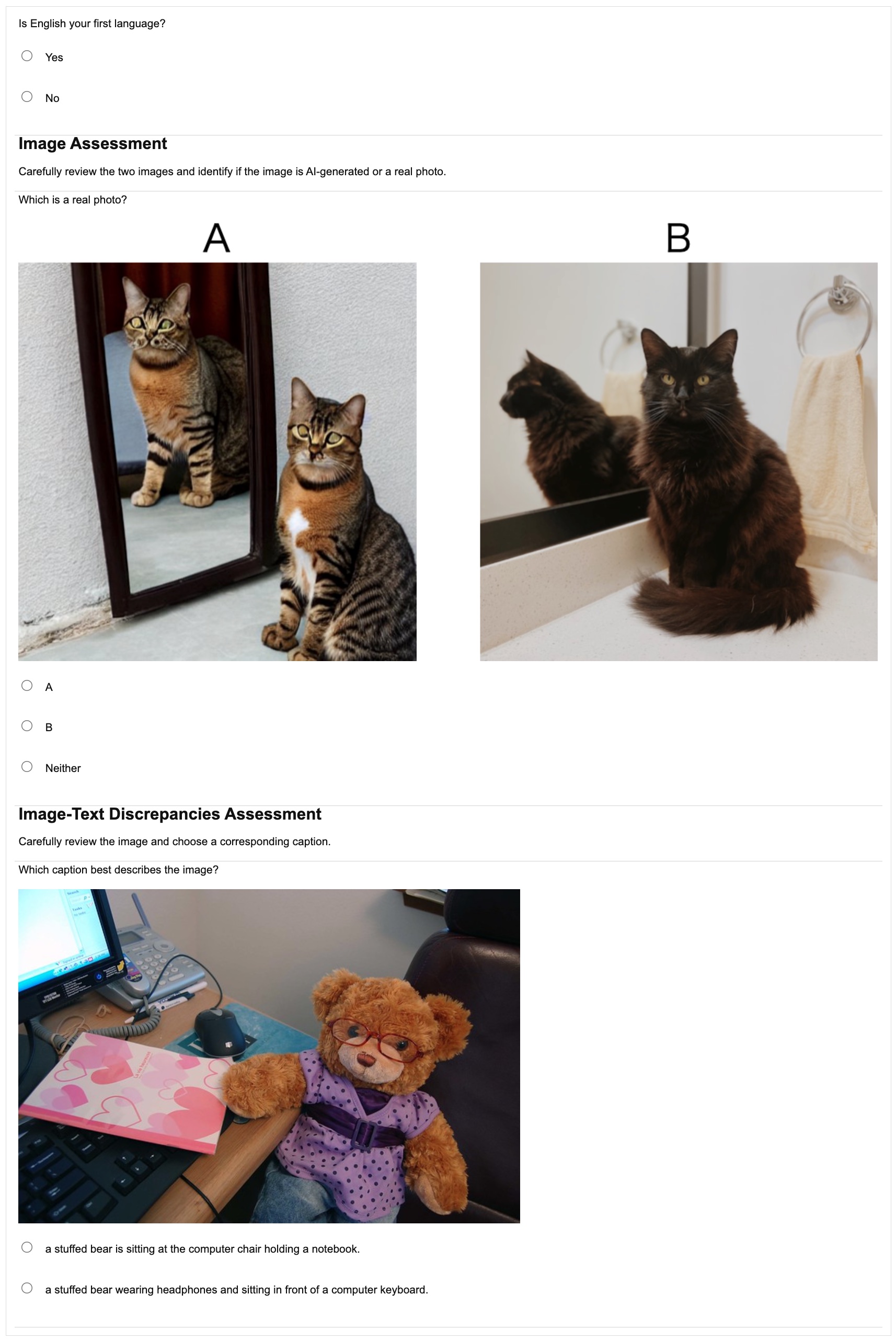}
    \caption{Screenshot of the pre-task qualification test for \textit{skillfulness} qualification. Annotators who answer that their first language is English and give correct answers to the two quiz are allowed to work for our annotation task.}
    \label{fig:skill_check}
\end{figure}

We show screenshot of our instructions for the annotation task (\cref{fig:instruction}), annotation interface (\cref{fig:anno_interface}), and pre-task qualification test for \textit{skillfulness} qualification (\cref{fig:skill_check}).
The implementation of the interfaces will be published.

\newpage
\section{Detailed results of human and automatic evaluation}
\begin{figure}[H]
    \centering
    \includegraphics[width=0.7\linewidth,clip]{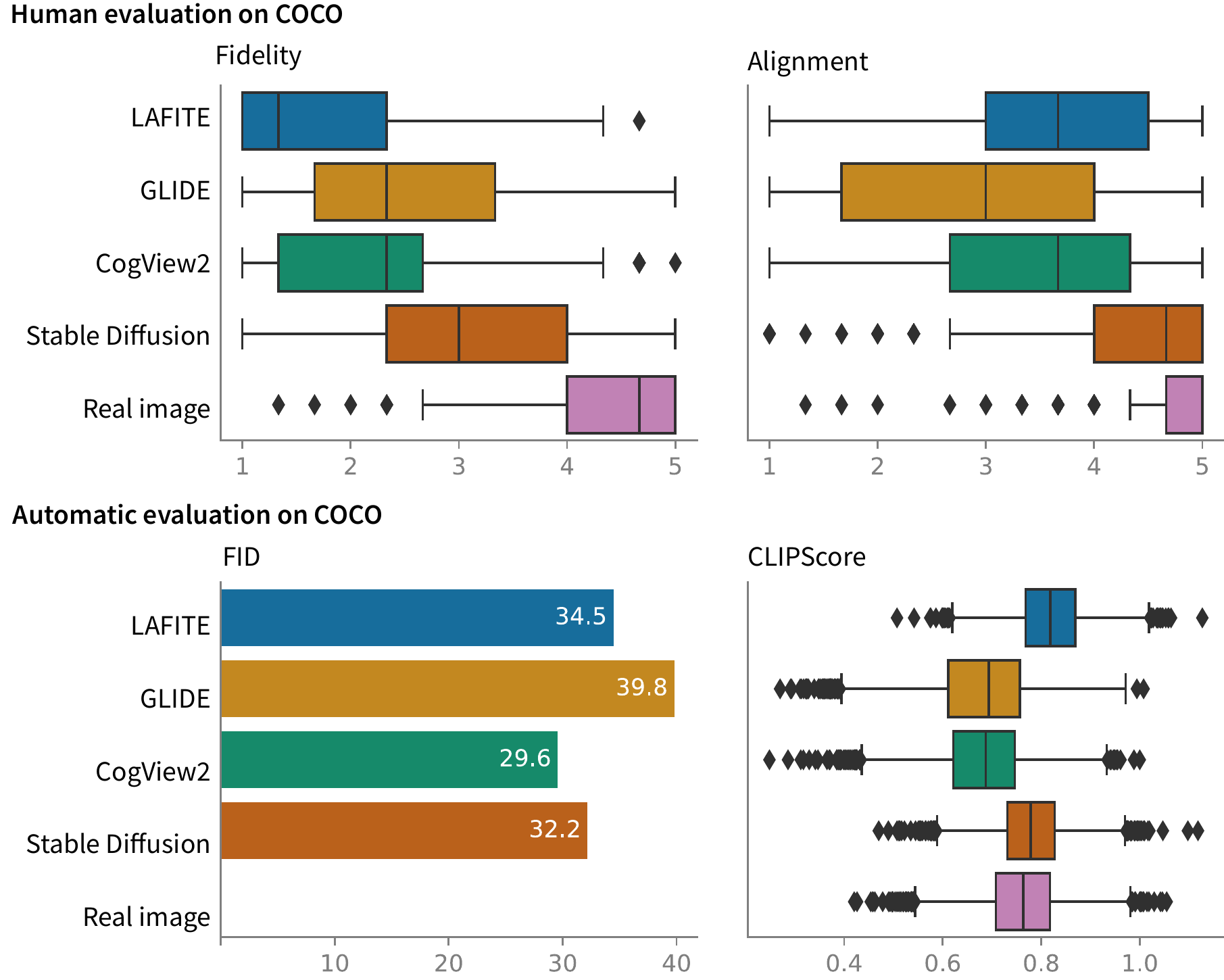}
    \caption{Evaluation results on COCO. (Top) Distributions of human ratings for fidelity and alignment. (Bottom) Automatic evaluation results. The bottom right plot shows the distribution of sample-level CLIPScore.}
    \label{fig:coco_result}
\end{figure}

\Cref{fig:coco_result} shows the human and automatic evaluation results on COCO.
The result demonstrates that the automatic measures do not align with human evaluation.
\Cref{tab:coco_test_result} shows the pairwise comparison results by a Tukey's HSD test.
We also compute Hedge's g values which indicate the differences between two methods are over one standard deviation apart.
We confirm that the ratings provided for each model show significant differences.

\begin{figure}[H]
    \centering
    \includegraphics[width=0.7\linewidth,clip]{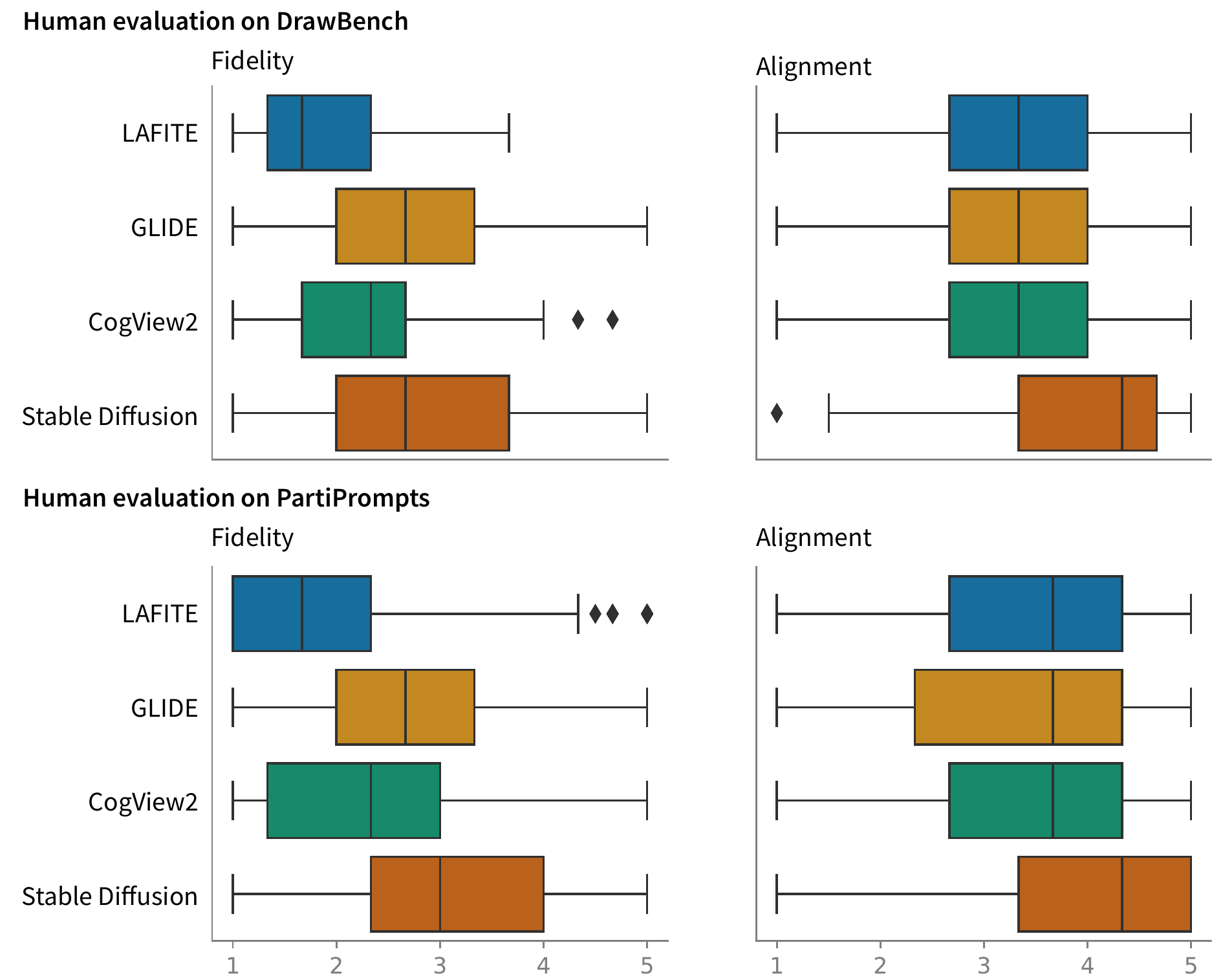}
    \caption{Distributions of human ratings for fidelity and alignment on DrawBench (top) and PartiPrompts (bottom).}
    \label{fig:drawbench_parti_result}
\end{figure}

On DrawBench, Stable Diffusion and GLIDE obtain comparable ratings for fidelity as shown in \cref{fig:drawbench_parti_result}.
For alignment, annotators rate Stable Diffusion the best, while the other models do not show significant differences as in \cref{tab:drawbench_test}.
On PartiPrompts, we observe similar trends, but the difference of fidelity ratings between Stable Diffusion and GLIDE is statistically significant as in \cref{tab:parti_test}. 

\begin{table}[H]
    \centering
    \caption{Pairwise post-hoc test with Tukey’s HSD test for ratings of Fidelity and Alignment on COCO. The numbers provided in the table are p-values, and the numbers in parentheses are effect size (Hedge's $g$).}
    \label{tab:coco_test_result}
    \begin{tabular}{rcccc}
    \toprule
     & Real image & CogView2 & GLIDE & LAFITE \\
    \midrule
    \multicolumn{4}{l}{\small Fidelity} \\
    CogView2 & 0.0000 (-2.79) & --- & --- & --- \\
    GLIDE & 0.0000 (-2.27) & 0.0000 (0.39) & --- & --- \\
    LAFITE & 0.0000 (-3.69) & 0.0000 (-0.48) & 0.0000 (-0.89) & --- \\
    Stable Diffusion & 0.0000 (-1.45) & 0.0000 (0.84) & 0.0000 (0.48) & 0.0000 (1.31)\\
    \midrule
    \multicolumn{4}{l}{\small Alignment} \\
    CogView2 & 0.0000 (-1.50) & --- & --- & --- \\
    GLIDE & 0.0000 (-1.85) & 0.0000 (-0.49) & --- & --- \\
    LAFITE & 0.0000 (-1.45) & 0.0002 (0.18) & 0.0000 (0.68) & --- \\
    Stable Diffusion & 0.0000 (-0.67) & 0.0000 (0.85) & 0.0000 (1.28) & 0.0000 (0.72) \\
    \bottomrule
    \end{tabular}
    
\end{table}

\begin{table}[H]
    \centering
    \caption{Pairwise post-hoc test with a Tukey’s HSD test for ratings of Fidelity and Alignment on DrawBench. The numbers provided in the table are p-values, and the numbers in parentheses are effect size (Hedge's $g$).}
    \label{tab:drawbench_test}
    \begin{tabular}{rccc}
    \toprule
     & CogView2 & GLIDE & LAFITE \\
    \midrule
    \multicolumn{4}{l}{\small Fidelity} \\
    GLIDE & 0.0000 (0.48) & --- & --- \\
    LAFITE & 0.0000 (-0.58) & 0.0000 (-1.06) & --- \\
    Stable Diffusion & 0.0000 (0.65) & 0.1034 (0.21) & 0.0000 (1.19)\\
    \midrule
    \multicolumn{4}{l}{\small Alignment} \\
    GLIDE & 0.5558 (0.13) & --- & --- \\
    LAFITE & 0.9352 (0.06) & 0.8884 (-0.07) & --- \\
    Stable Diffusion & 0.0000 (0.71) & 0.0000 (0.64) & 0.0000 (0.70)\\
    \bottomrule
    \end{tabular}
\end{table}

\begin{table}[H]
    \centering
    \caption{Pairwise post-hoc test with Tukey’s HSD test for ratings of Fidelity and Alignment on PartiPrompts. The numbers provided in the table are p-values, and the numbers in parentheses are effect size (Hedge's $g$).}
    \label{tab:parti_test}
    % Fidelity
    \begin{tabular}{rccc}
    \toprule
     & CogView2 & GLIDE & LAFITE \\
    \midrule
    \multicolumn{4}{l}{\small Fidelity} \\
    GLIDE & 0.0000 (0.34) & --- & --- \\
    LAFITE & 0.0000 (-0.37) & 0.0000 (-0.73) & --- \\
    Stable Diffusion & 0.0000 (0.67) & 0.0000 (0.33) & 0.0000 (1.07) \\
    \midrule
    \multicolumn{4}{l}{\small Alignment} \\
    GLIDE & 0.1371 (-0.07) & --- & --- \\
    LAFITE & 0.9504 (0.02) & 0.0363 (0.10) & --- \\
    Stable Diffusion & 0.0000 (0.57) & 0.0000 (0.62) & 0.0000 (0.58)\\
    \bottomrule
    \end{tabular}
\end{table}

\section{Captions and images used for sample size analysis}
\begin{figure}[H]
    \centering
    \includegraphics[width=0.9\linewidth,clip]{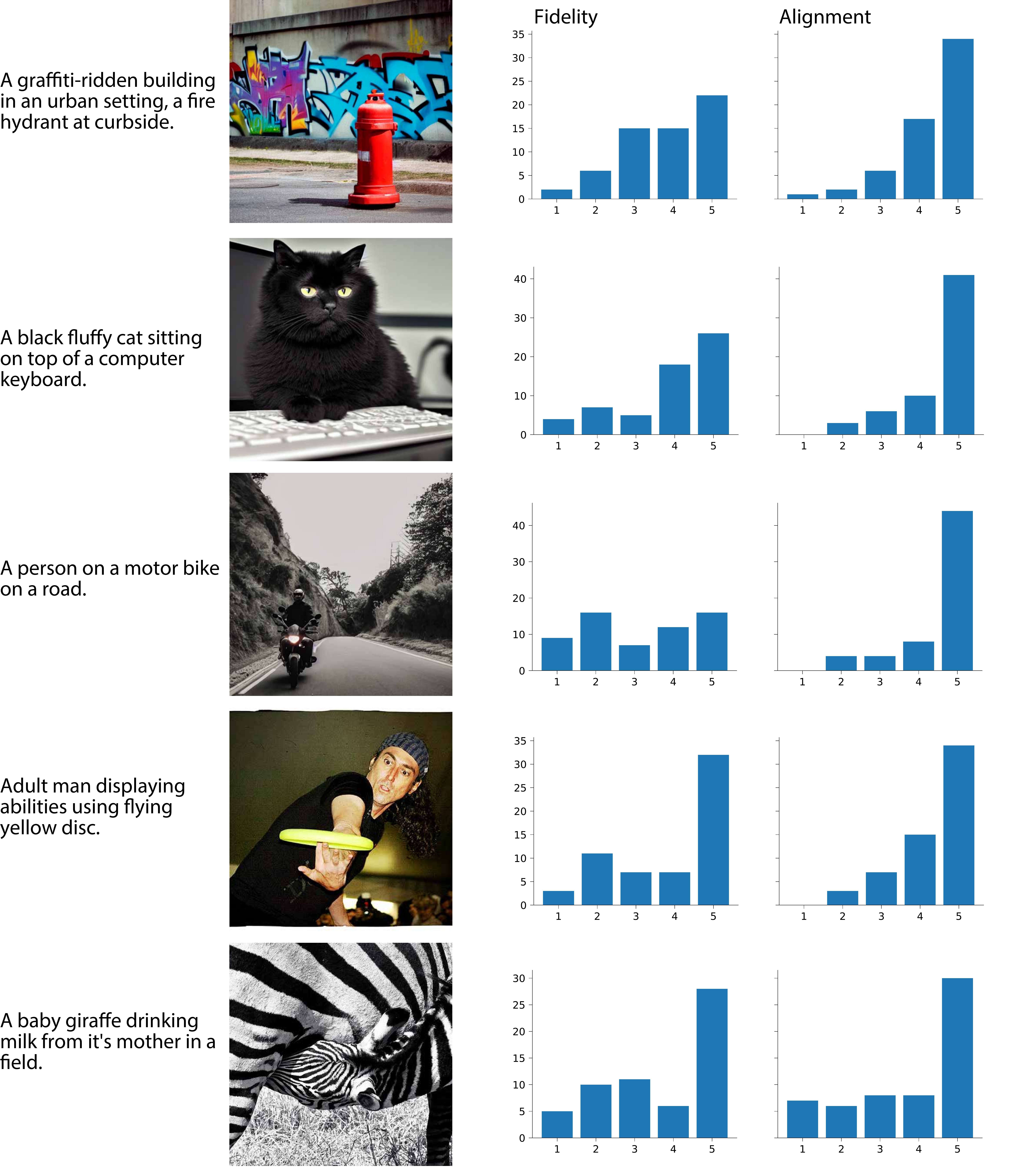}
    \caption{Image and caption pairs used for annotator size analysis. Histograms represent distributions of human ratings.}
    \label{fig:anno_size_samples_a}
\end{figure}

\begin{figure}[H]
    \centering
    \includegraphics[width=0.9\linewidth,clip]{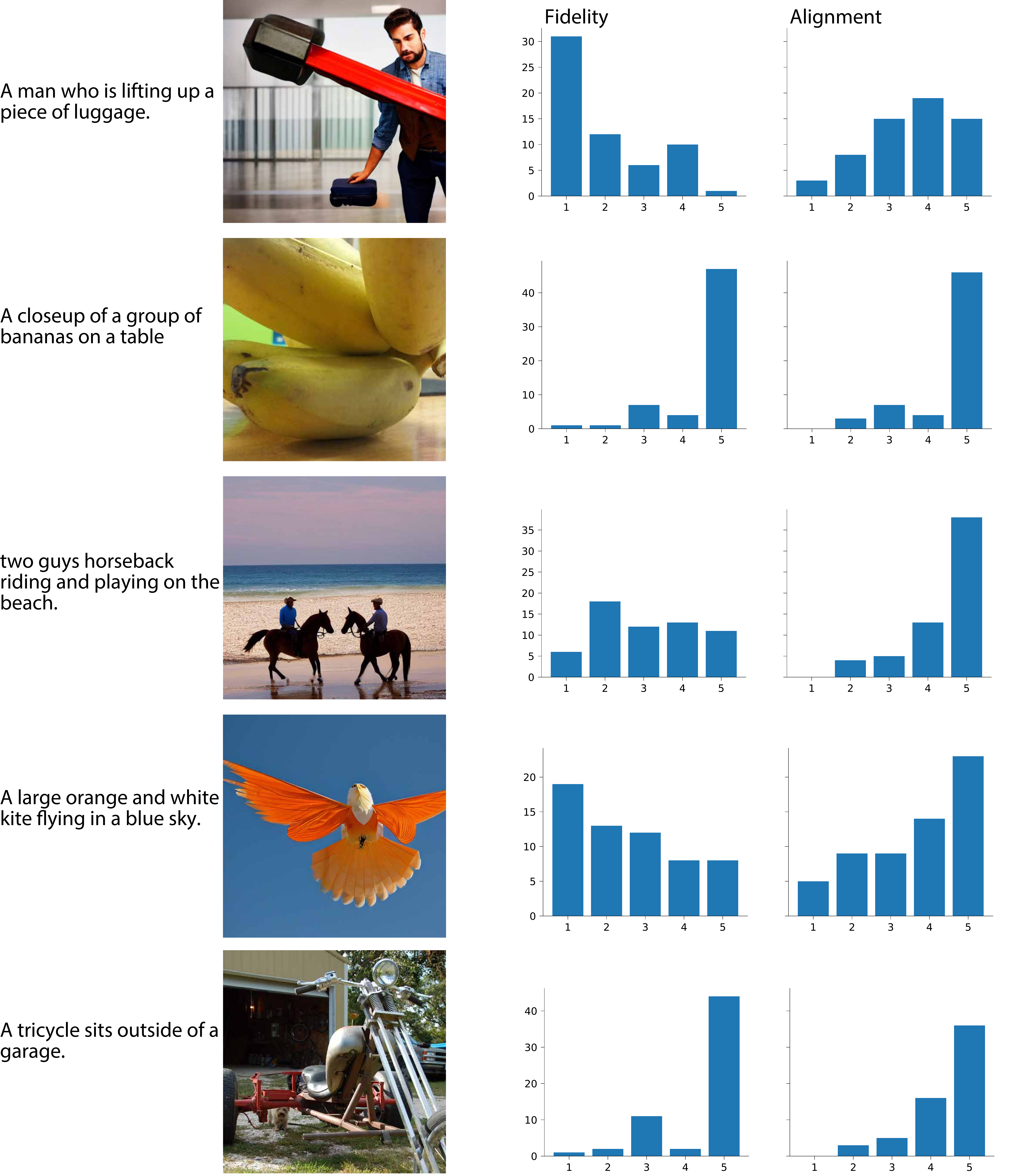}
    \caption{Image and caption pairs used for annotator size analysis. Histograms represent distributions of human ratings.}
    \label{fig:anno_size_samples_b}
\end{figure}

\begin{figure}[H]
    \centering
    \includegraphics[width=0.9\linewidth,clip]{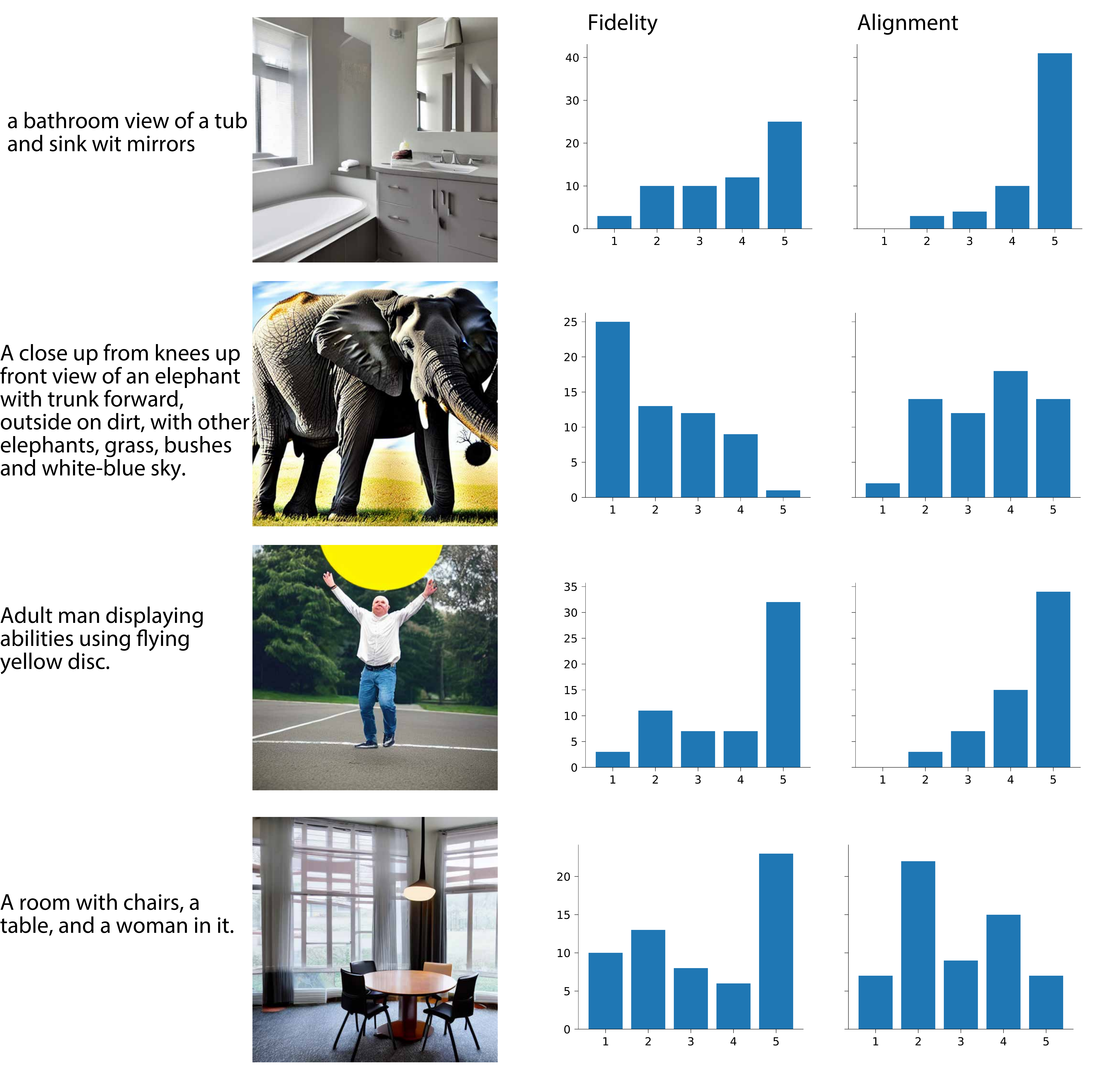}
    \caption{Image and caption pairs used for annotator size analysis. Histograms represent distributions of human ratings.}
    \label{fig:anno_size_samples_c}
\end{figure}

The caption and image pairs used for annotation size analysis are shown in \cref{fig:anno_size_samples_a,fig:anno_size_samples_b,fig:anno_size_samples_c}.
We selected samples where three annotators gave diverse labels in a pilot data collection. 
The histogram plots show distributions of human ratings by 60 annotators.
We observe variations in the ratings, and some samples show several peaks.
This observation indicates that human evaluation measures other than averaging ratings may be needed.

\section{Reporting experimental details for transparency}
Our literature review revealed that many papers omit details of experimental configurations of human evaluation.
To the alleviate transparency issue, we offer templates for reporting human evaluation settings.
\Cref{tab:example_details} summarizes recommended details to report.
For customizable sample text, see \Cref{fig:smpl_text}.

\begin{table}[H]
\caption{Example report of a human evaluation setting.}
\label{tab:example_details}
\centering
\begin{tabular}{@{}rl@{}}
\toprule
\multicolumn{2}{l}{Dataset details}                                                                \\ \midrule
\#captions              & 1000                                                                     \\
\#ratings / item        & 3                                                                        \\
\#unique annotators     & 148                                                                      \\
Tested models           & LAFITE, GLIDE, CogView2, Stable Diffusion, Real image                    \\
Types of rating         & 5-point Likert scale                                                     \\
Evaluation criteria     & Fidelity, Alignment to caption                                           \\ \midrule
\multicolumn{2}{l}{Annotation details}                                                             \\ \midrule
Platform                & AMT                                                                      \\
Annotator qualification & i) Over 18 years old and agreed to work with potentially offensive content. \\
                        & ii) AMT Masters                                                              \\
Compensation            & \$0.05 / task                                                            \\
Interface               & \Cref{fig:anno_interface}                                                                         \\
Instructions            & \Cref{fig:instruction}                                                                         \\ 
IAA &  Fidelity: 0.41, Alignment: 0.48 (Krippendorff's $\alpha$)\\
\bottomrule
\end{tabular}
\end{table}

\begin{figure}[H]
We collected annotations for images generated by \fbox{tested models} for \fbox{\#captions} captions, resulting in \fbox{\#annotations in total} annotations.
Annotators are invited on \fbox{crowdsourcing platform}.
Annotators who \fbox{annotator qualification} are allowed to work on our task.
Krippendorff's $\alpha$ is \fbox{$\alpha$ value}.
\fbox{\#unique annotators} annotators participated in total, and the average number of tasks per annotator was \fbox{average \#tasks per annotator}.
Annotators get \fbox{compensation per task} for each instance of the task
The median time spent on one task is \fbox{time spent per task}; that is, the expected hourly wage is \fbox{expected hourly wage}.
    \caption{Sample template for reporting human evaluation settings.}
    \label{fig:smpl_text}
\end{figure}

\section{Potential Bias in Annotator Ratings}
Even with carefully designed instructions for annotators, they may annotate a text-image pair differently.
The causes of the disagreement between ratings of multiple annotators are, for example, the subjectivity of annotation tasks and the different stringency of annotators.
We therefore describe a potential concern of annotator biases in datasets collected through crowdsourcing.

\begin{figure}[H]
    \centering
    \includegraphics[width=0.8\linewidth,clip]{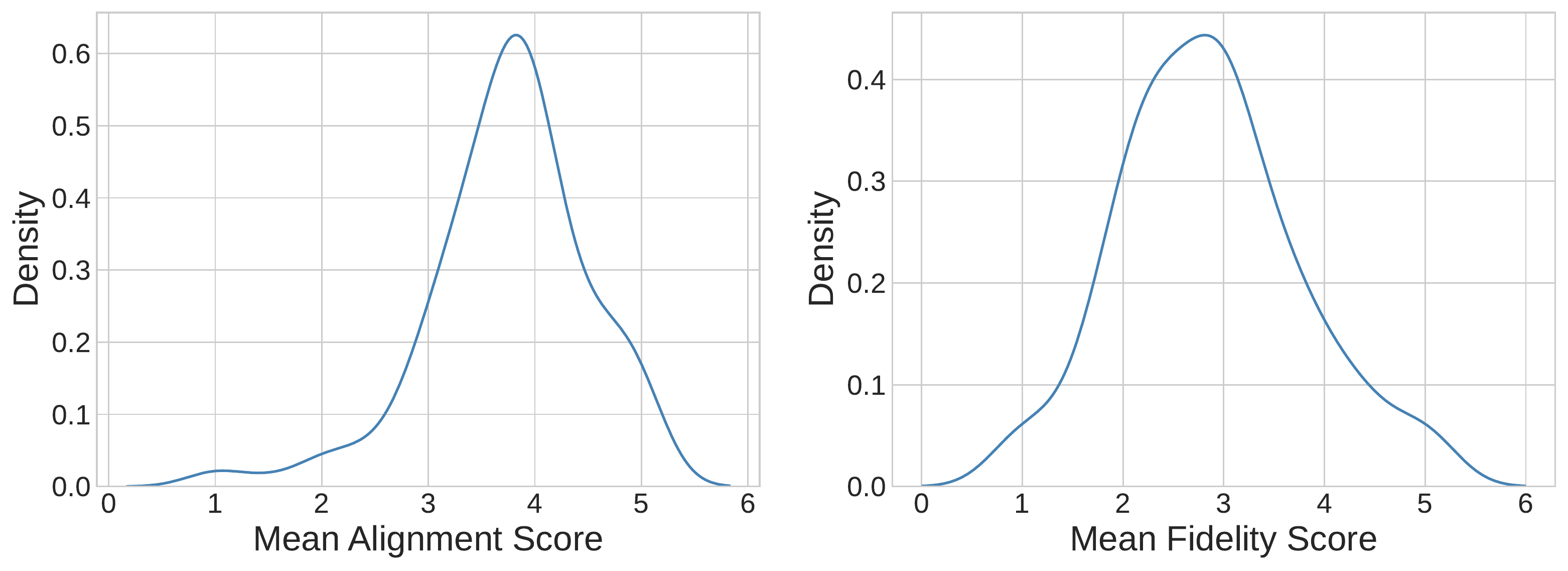}
    \caption{Distributions of mean ratings of each annotator for fidelity and alignment.}
    \label{fig:distribution_mean_rating}
\end{figure}

\begin{figure}[H]
    \centering
    \includegraphics[width=0.8\linewidth,clip]{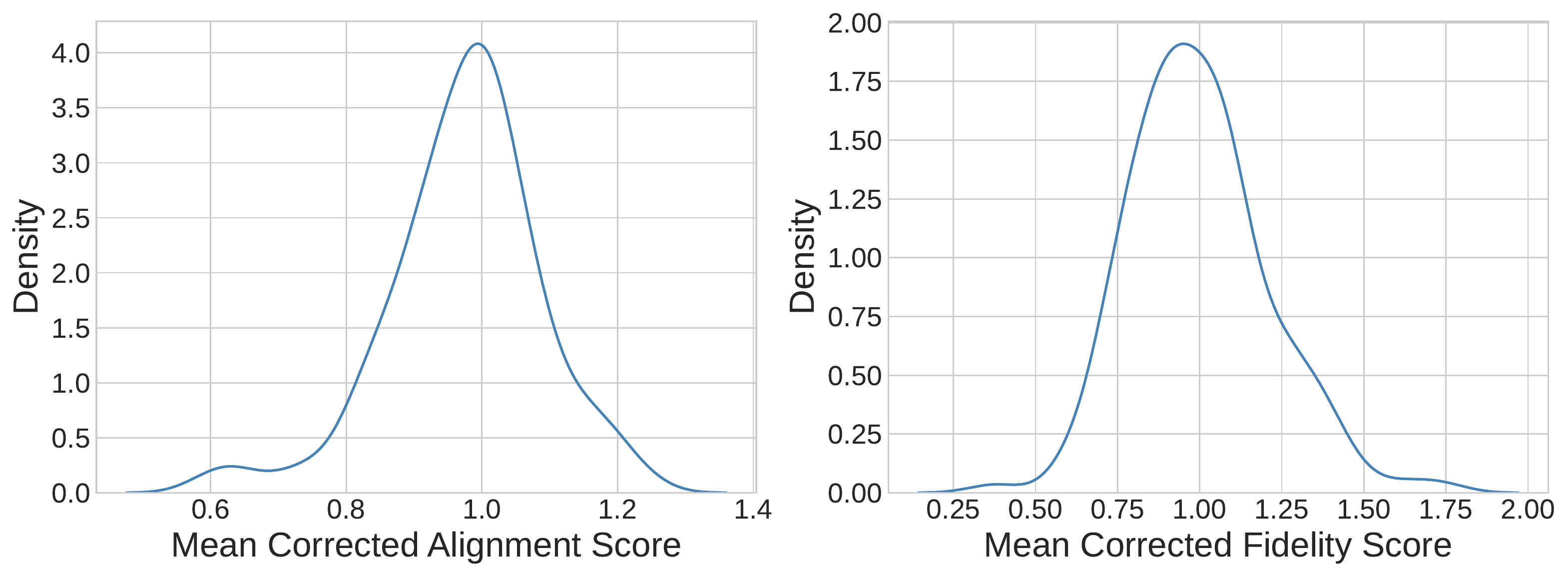}
    \caption{Distributions of mean corrected ratings of each annotator for fidelity and alignment.}
    \label{fig:distribution_corrected_mean_rating}
\end{figure}

Each plot in \Cref{fig:distribution_mean_rating} shows the distribution of the mean ratings of the annotators in our collected dataset.
The x- and y-axes indicate the mean rating score for each annotator and the density of annotators, respectively.
The mean alignment/fidelity scores are computed by taking the average of one annotator's ratings for different samples.
We observe a non-negligible number of annotators who provide highly biased ratings (the right and left tails of distributions).
This result suggests that there may exist annotator-dependent rating biases.
However, this can also happen when the tasks assigned to each annotator have imbalanced true ratings; in this case, annotators with biased mean ratings may correctly judge their tasks.
We consider a rating correction strategy for task-dependent rating biases to remove the effect of unbalanced task assignments.
We first compute the mean ratings for each task (\ie, text-image pair) and then normalize each rating using the mean.
\Cref{fig:distribution_corrected_mean_rating} demonstrates the distributions of mean corrected ratings for fidelity and alignment.
It can be observed that the distributions are more ``centered'', and the density of annotators with extreme mean scores is reduced compared to that without the rating correction.
On the other hand, there are a few annotators with biased (corrected) ratings, particularly in mean alignment scores.

In summary, the annotator-dependent bias in quality ratings is not severe in our collected dataset; this is also confirmed by the high Krippendorff's $\alpha$ values.
We may further enhance the data quality by considering the annotator-dependent bias in the aggregation of multiple ratings to generate reliable ground truth labels.

%%%%%%%%% REFERENCES
\newpage
{\small
\bibliographystyle{ieee_fullname}
\bibliography{egbib}
}